\documentclass[a4paper,fleqn]{cas-dc}

\def\tsc#1{\csdef{#1}{\textsc{\lowercase{#1}}\xspace}}
\tsc{WGM}
\tsc{QE}
\tsc{EP}
\tsc{PMS}
\tsc{BEC}
\tsc{DE}

\usepackage{graphicx}  
\usepackage{caption}
\usepackage{float}
\usepackage{stfloats}
\usepackage{amsmath}
\usepackage{booktabs}
\usepackage{pifont}
\usepackage{soul}
\usepackage[numbers,sort&compress]{natbib}
\usepackage{multirow}
\usepackage{subfigure} 
\begin{document}
\let\WriteBookmarks\relax
\def\floatpagepagefraction{1}
\def\textpagefraction{.001}
\let\printorcid\relax   
\shorttitle{} 
\setlength{\abovedisplayskip}{2pt}


\title [mode = title]{RFAConv: Receptive-Field Attention Convolution for Improving Convolutional Neural
	Networks}                      



%
\author[1]{Xin Zhang}
\author[1]{Chen Liu}
\author[1]{Tingting Song}
\cormark[1]
\ead{E-mail address: ttsong@cqnu.edu.cn}
\author[1,2]{Degang Yang }
\cormark[1]
\ead{E-mail address: yangdg@cqnu.edu.cn}
\author[3]{Yichen Ye }
\author[1]{Ke Li}
\author[1]{ Yingze Song}






\affiliation[1]{organization={College of Computer and Information Science, Chongqing Normal University},
	city={Chongqing},
	postcode={401331}, 
	country={china}}



\affiliation[2]{organization={Chongqing Engineering Research Center of Educational Big Data Intelligent Perception and Application},
  city={Chongqing},
  postcode={401331}, 
  country={china}}

\affiliation[3]{organization={College of Electronic and Information Engineering, Southwest University},
	city={Chongqing},
	postcode={400715}, 
	country={china}}

\cortext[cor1]{Corresponding author.}




\begin{abstract}
In the realm of deep learning, spatial attention mechanisms have emerged as a vital method for enhancing the performance of convolutional neural networks. However, these mechanisms possess inherent limitations that cannot be overlooked. This work delves into the mechanism of spatial attention and reveals a new insight. It is that the mechanism essentially addresses the issue of convolutional parameter sharing. By addressing this issue, the convolutional kernel can efficiently extract features by employing varying weights at distinct locations. However, current spatial attention mechanisms focus on shallow attention to spatial features, which is insufficient to address the fundamental challenge of parameter sharing in convolutions involving larger kernels. In response to this challenge, we introduce a novel attention mechanism known as Receptive-Field Attention (RFA). Compared to existing spatial attention methods, RFA not only concentrates on the receptive-field spatial features but also offers effective attention weights for large convolutional kernels. Building upon the RFA concept, a Receptive-Field Attention Convolution (RFAConv) is proposed to supplant the conventional standard convolution. Notably, it offers nearly negligible increment of computational overhead and parameters, while significantly improving network performance. Furthermore, this work reveals that current spatial attention mechanisms require enhanced prioritization of receptive-field spatial features to optimize network performance. To validate the advantages of the proposed methods, we conduct many experiments across several authoritative datasets, including ImageNet, COCO, VOC, and Roboflow. The results demonstrate that the proposed methods bring about significant advancements in tasks, such as image classification, object detection, and semantic segmentation, surpassing convolutional operations constructed using current spatial attention mechanisms. Presently, the code and pre-trained models for the associated tasks have been made publicly available at \url{https://github.com/Liuchen1997/RFAConv}.
\end{abstract}


\begin{highlights}
\item In this work, the spatial attention mechanism is analyzed from a new perspective.

\item A novel attention mechanism is proposed, which solves the problem of convolution kernel parameter sharing.

\item  RFAConv is proposed which can be considered as a high performance method to replace the $3~\times~3$ standard convolution.

\item RFCBAM and RFCA are integrated with standard convolution to develop non-shared parameter convolutional operations.

\end{highlights}


\maketitle

\section{Introduction}

Convolutional neural networks have dramatically reduced the computational overhead and complexity of models by using the convolutional operation with shared parameters. Currently, convolutional neural networks have now established a complete system and formed advanced convolutional neural networks, such as DCANet \cite{BAI2025111379}, HDTCNet\cite{GU2025111837} and FSENet \cite{HU2025111425}.

\begin{figure}[h]
	\centering
	\includegraphics[trim=0 0 0 0,clip,scale=0.9]{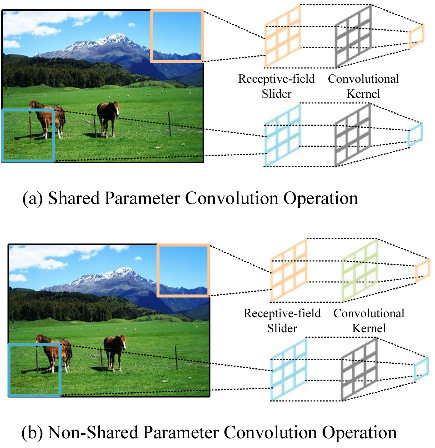}
	\caption{(a) Illustrates the process of extracting features from different locations using shared parameter convolution. (b) Illustrates the process of extracting features from different locations using non-shared parameter convolution. It is evident that information at different locations in the image varies, and using the same parameters to extract features does not fully consider the differentiation across different locations.}
	\label{shared-shown}
\end{figure}

However, standard convolution operations share parameters in each receptive-field slider, which limits network performance.  Elsayed et al. \cite{elsayed2022} asserted that applying the same weights at all locations may be a less efficient use of computation than applying different weights at different locations. Jin et al. \cite{jin2022lagconv} argued that the standard convolution operation suffers from context-independence and that it should be fully aware of the specificity of each pixel. After carefully studying these works and standard convolutional operations, we gained inspiration. For the standard convolution operation, it uses the same parameters in each receptive-field to extract information, which does not take into account the difference information at different locations, as shown in Figure~\ref{shared-shown} (a). Therefore, the aim of this work is to consider the design of a non-shared parameter convolutional operation to perceive the difference in information at different locations, which is mainly achieved by assigning different convolutional parameters to the convolutional kernel at different locations, as depicted in Figure~\ref{shared-shown} (b) for clarity.

Building upon the aforementioned goals, we examine the inherent relationship between spatial attention mechanisms and conventional convolutions from a fresh standpoint. Upon analysis, we discover that the current spatial attention mechanism  fundamentally address the issue of parameter sharing in convolutional operations, but it remains restricted to the recognition of spatial features. The current spatial attention mechanism does not fully address the parameter sharing problem for larger convolutional kernels. Furthermore, they are unable to emphasize the significance of each feature in the receptive-field, such as the existing Convolutional Block Attention Module (CBAM) \cite{woo2018cbam} and Coordinate Attention (CA) \cite{hou2021coordinate}.

In response to the aforementioned challenges, we introduce the innovative Receptive-Field Attention (RFA) mechanism, which not only effectively addresses the parameter sharing issue in convolution kernels but also meticulously considers the significance of each feature within the receptive-field, enabling precise feature capture and efficient utilization. Building upon the RFA concept, we introduce the Receptive-Field Attention Convolution (RFAConv) as a replacement for traditional standard convolution operations. With minimal additional parameters and computational overhead, RFAConv substantially enhances network performance, potentially supplanting traditional convolutional operations as a fundamental component in neural network architectures. Extensive experiments conducted on authoritative datasets like ImageNet, COCO, VOC, and Roboflow for tasks such as classification, object detection, and semantic segmentation, demonstrate that RFAConv yields significant improvements in image classification and object detection compared to convolutional operations utilizing existing spatial attention mechanisms. Moreover, we assert that the future advancement of spatial attention mechanisms should heavily emphasize receptive-field spatial features. By delving deep into this area's potential, the strengths of convolutional neural networks will be further amplified. To substantiate this claim, the existing attention mechanism is optimized to seamlessly integrate receptive-field spatial features, effectively resolving the convolutional kernel parameter sharing issue. Therefore, building upon the foundational architectures of CBAM and CA, we propose two enhanced attention mechanisms RFCBAM and RFCA, and conduct comprehensive comparative experiments across various tasks, including classification, target detection, and semantic segmentation. The results strongly endorse our perspective, showcasing that attention mechanisms focusing on receptive-field spatial features can significantly enhance network performance once more.

\section{Related Works}
\subsection{Convolutional neural network architecture}
After conducting a thorough study on convolutional operations, Dai et al. \cite{dai2017deformable} claimed that the convolutional operation with a fixed sampling position can restrict the performance of the network to a certain extent. So, they proposed Deformable Conv, which alters the sampling positions of convolutional kernels by learning offsets. The convolutional operation, which serves as a basic operation in convolutional neural networks, has led the development of many advanced network mode. He et al. \cite{he2016deep} suggested that as the depth of the network increases, the model becomes harder to train and may experience a degradation phenomenon. To address this issue, they proposed to use residual connections to revolutionize the design of network.
Chen et al. \cite{chen2023coupled} proposed the Coupled Global–Local (CGL) network for enhancing thee efficiency of object detection in large aerial images. It shares information from global branches and local branches, and can adaptively adjust the receptive-field for better feature extraction. Li et al. \cite{li2023large} devised the Large Selective Kernel Network (LSKNet), which can dynamically adjust its large spatial receptive-field to better model the ranging context of various objects in remote sensing scenarios. Ma et al. \cite{ma2018shufflenet} discovered that models with few parameters do not always result in faster inference times, and similarly, small computational overhead does not guarantee quick performance. After careful study, they proposed the ShuffleNet V2. The YOLO object detection network divided the input image into a grid to predict the location and class of objects. As research has progressed, eight versions of object detectors based on YOLO have been proposed, such as YOLOv5 \cite{YOLOV5}, YOLOv7 \cite{wang2023yolov7}, YOLOv8 \cite{YOLOV8}, etc. While the previously-mentioned convolutional neural network architectures have achieved significant success, they do not directly address the problem of parameter sharing during the feature extraction process. Our work focuses on utilizing the attention mechanism to tackle the problem of convolutional parameter sharing from a fresh perspective.

\subsection{Attention Mechanism}
Attention mechanism, as a technique to improve the performance of network models, allows models to focus on key features. The theory of attention mechanism has now established a complete and mature system in the field of deep learning.  Wang et al. \cite{wang2020eca} asserted that the correspondence between individual channels and weights is indirect when the Squeeze-and-Excitation Block interacts with information. Therefore, they designed the Efficient Channel Attention (ECA) by replacing the Fully Connected (FC) layer in the SE with a one-dimensional convolution of adaptive kernel size. Woo et al. \cite{woo2018cbam} proposed the Convolutional Block Attention Module (CBAM), which combines channel attention and spatial attention. As a plug-and-play module, it can be embedded into convolutional neural networks to enhance network performance. Although SE and CBAM have allowed the network to achieve good performance, Hou et al. \cite{hou2021coordinate} found that the compressing feature in SE and CBAM lost too much information. Therefore, they proposed the lightweight coordinate attention (CA) to solve this problem. Zhang et al. \cite{zhang2022epsanet} generated feature maps at different scales on channels to build a more efficient channel attention mechanism. Misra et al. \cite{Misra_2021_WACV} proposed TripletAttention (TA), which de-transforms the dimensions of features and obtains three attention weights through three different branches. Finally, the branches are fused by weighting and averaging to obtain efficient feature. Yu et al. \cite{yu2023mca} argued that most existing approaches to attention mechanisms either ignore modelling attention in the channel and spatial dimensions, or introduce higher model complexity and greater computational burden. To alleviate this dilemma, they proposed a lightweight and efficient multidimensional collaborative attention (MCA), which simultaneously infers attention in the channel, height and width dimensions by using a three-branch architecture. Li et al. \cite{li2022ham} introduced Hybrid Attention Module (HAM), which employs a channel attention module to simultaneously construct channel attention maps and produce channel-refined feature representations. Building upon these channel-wise attention weights, the spatial attention module subsequently partitions the feature maps along the channel dimension, generating complementary spatial attention descriptors through dual-path processing. These descriptors are then systematically integrated through attention-weighted feature modulation, enabling adaptive refinement of spatial feature representations.

This work introduces a new approach to address the issue of parameter sharing in standard convolutional operations. Our proposal involves combining attention mechanisms to create convolutional operations. Although existing attention mechanisms have demonstrated good performance, they do not specifically target the spatial features of receptive-fields. To tackle this limitation, we developed RFAConv with non-shared parameters to improve performance of the network.

\section{Methods}
Existing works predominantly concentrate on leveraging spatial attention to enhance network performance, exploring its contributions to feature extraction, information screening. While they indeed show the effectiveness of spatial attention to some degree, these analyses often approach spatial attention from a conventional viewpoint, overlooking the deeper relationship between its intrinsic mechanism and convolutional kernels. Therefore, this section  analyses the intrinsic connection between spatial attention mechanisms and standard convolutional operations from a new perspective, and points out that existing spatial attention mechanisms do not adequately address the problem of parameter sharing for large convolutional kernels. To address this issue, we propose the concept of the receptive-field spatial feature, in which different attention weights can be learned for each receptive-filed slider. Based on this, we introduce a receptive-field attention convolution (RFAConv), which can be primarily segmented into Group Optimization and Receptive-Field Attention Process stages. Following these two processing steps, RFAConv effectively resolves the issue of parameter sharing to enhance network performance in classification, target detection, and semantic segmentation tasks.

\subsection{Spatial Attention and Standard Convolutional Operations}
As is widely recognized, incorporating attention mechanisms into convolutional neural networks can enhance their performance. Regarding the effectiveness of spatial attention mechanisms, a new perspective is analyzed in this work.
It effectively overcomes the inherent limitation of convolutional neural networks, which is parameter sharing. Currently, the most commonly used kernel sizes in convolutional neural networks are 1 $\times$ 1 and 3 $\times$ 3.  The convolutional operation for extracting features after introducing the spatial attention mechanism is either a 1 $\times$ 1 or a 3 $\times$ 3 convolutional operation. To intuitively show the process, the spatial attention mechanism is inserted into the front of the 1  $\times$ 1 convolutional operation. The input feature map is weighted (Re-weight "$\times$") by the attention map. Finally, the slider feature information of the receptive-field is extracted through 1 $\times$ 1 convolution operation. The overall process can be simply represented as the Equation~(\ref{qe2}). Due to the 1 $\times$ 1 convolution operation has only one convolution parameter, $K$ still represents only one parameter value, $A_{i}$ denotes a attention weight for the corresponding position,  $F_{i}$ represents the value obtained by each convolutional slider after computation and N still represents the total number of the receptive-field slider. Unlike the 3 $\times$ 3 convolution, the 1 $\times$ 1 convolution has just one pixel value $X_{i}$ in each receptive-field slider.

\begin{equation}
\small
\begin{aligned}
&F_{1} =  X_{1} \times A_{1} \times K \\
&F_{2} =  X_{2} \times A_{2} \times K \\
&...\\
&F_{N} =  X_{N} \times A_{N} \times K  \\
\label{qe2}
\end{aligned}
\end{equation}
In the absence of spatial attention, each sliding window obtains $F_{i}$ simply by performing operations on $X_{i}$ with $K$. Therefore, analogous to standard convolution operations, if we take spatial attention weighting and convolution operations as an overall arithmetic step, the interesting thing is that the problem of parameter sharing in the extraction of features by 1 $\times$ 1 convolution operations is solved. In other words, the value of $ A_{i}\times K $ is treated as a new convolution kernel parameter.

\begin{figure*}
\begin{equation}
\label{eq3}
\begin{aligned}
&F_{1} =  X_{11} \times A_{11}  \times K_{1} +  X_{12}\times A_{12} \times K_{2} + X_{13} \times A_{13} \times K_{3} +...+ X_{19} \times A_{19} \times K_{9} \\
&F_{2} =  X_{21} \times A_{21} \times K_{1} +  X_{22} \times A_{22} \times K_{2} + X_{23} \times A_{23} \times K_{3} +...+ X_{29} \times A_{29} \times K_{9} \\
&...\\
&F_{N} =  X_{N1} \times A_{N1} \times K_{1} +  X_{N2} \times A_{N2} \times K_{2} + X_{N3} \times A_{N3} \times K_{3} +...+ X_{N9} \times A_{N9} \times K_{9} \\
\end{aligned}
\end{equation}
\end{figure*}

\begin{figure}[h]
	\centering
	\includegraphics[trim=0 0 0 0,clip,scale=0.75]{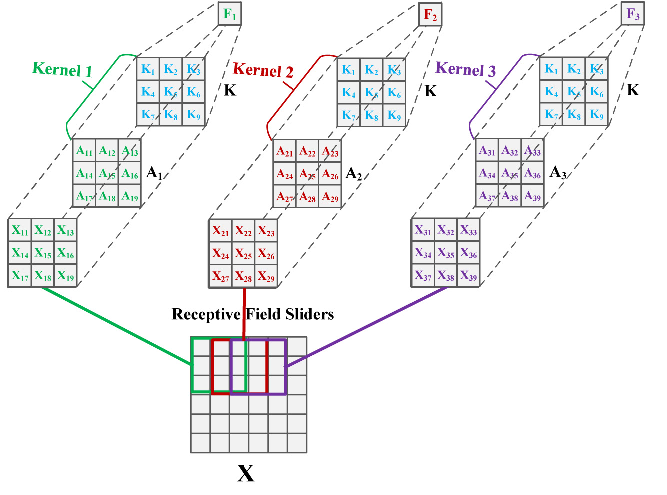}
	\caption{It is obvious that there is an overlap of features in each receptive-field slider, which leads to the problem of sharing of attentional weights across sliders.}
	\label{convolution_attention_3}
\end{figure}

Here $X_{ij}$ and $A_{ij}$ represent the $j$-th feature value and $j$-th attentional weight in the $i$-th receptive-field slider. $K_{i}$ represents the $i$-th convolution parameter in the convolution kernel $K$. However, when the spatial attention mechanism is inserted in front of the 3 $\times$ 3 convolutional operation, it will be limited. As mentioned above, if we take the value of $ A_{i} \times K $ as a new convolution kernel parameter, the Equation~(\ref{eq3}) completely solves the problem of parameter sharing for large-scale convolutional kernels. As shown in the Figure~\ref{convolution_attention_3}, it represents the process in which the input features are weighted by spatial attention and then operated by a 3 $\times$ 3 convolution. If the product of the attention weight corresponding to the receptive-field slider and the convolution kernel is taken as the convolution parameter, then the problem of convolution parameter sharing is solved. 

However, the most important point is that the convolutional kernel will share some of the features when it extracts features in each receptive-field slider. In other words, there will be an overlap within each receptive-field slider and when spatial attention is weighted, the size of the attention map is the same as the size of the input features, so the attention weights are shared in the receptive-field slider. Therefore, in Figure~\ref{convolution_attention_3}, there exists the case where $A_{12}=A_{21}, A_{13}=A_{22}, A_{15}=A_{24},~...~, A_{29}=A_{38}$. In this case, the weights of the spatial attention map are shared across each sliding window. As a result, the spatial attention mechanism cannot effectively address the problem of parameter sharing for large-scale convolutional kernels, such as the 3 $\times$ 3 convolution,  because it does not consider the receptive-field spatial features. Consequently, the effectiveness of the spatial attention mechanism is limited.

\subsection{Receptive-Filed Spatial Feature}
\begin{figure}[h]
	\centering
	\includegraphics[trim=0 0 0 0,clip,scale=1]{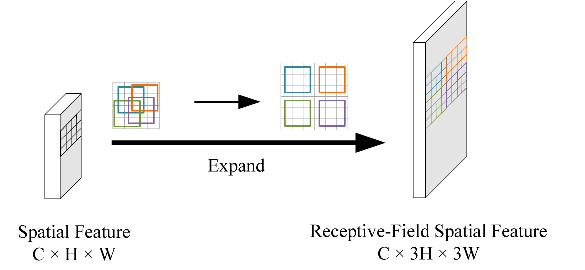}
	\caption{The receptive-field spatial features are obtained by expanding the spatial features.}
	\label{receptvie_field_spatial_feature}
\end{figure}

As highlighted in the preceding subsection, current spatial attention mechanisms share a fraction of their attention weights across the receptive-field slider, resulting in a lack of effective addressing of the convolutional parameter sharing issue. To tackle this challenge, our approach centers on the receptive-field spatial features. It is specifically designed for convolutional kernels in this work and is dynamically generated based on the kernel size. As shown in Figure~\ref{receptvie_field_spatial_feature}, the 3 $\times$ 3 convolutional kernel is used as an example.

In Figure~\ref{receptvie_field_spatial_feature}, the ``Spatial Feature" refers to the original feature map. The ``Receptive-Field Spatial Feature" is the feature map expanded by spatial features, which is composed of non-overlapping sliding windows. This is why the receptive-field spatial feature become three times wider and three times higher. In a subsequent section, we will describe in detail the method of expanding spatial features into receptive-field spatial features. Therefore, each 3 $\times$ 3 size window in the receptive-field spatial feature represents a receptive-field slider and is a non-overlapping sliding window. In this way, the weight of attention learned by focusing on the receptive-field spatial feature can not be shared in each receptive-field slider. Therefore after obtaining the attention weights and constructing a new operator with the standard convolution operation, then the problem of sharing the parameters of the standard convolution operation will be solved. Moreover, the spatial attention mechanism can solve the problem of convolutional parameter sharing by focusing on the receptive-field spatial feature, which is the direction for upgrading the existing spatial attention.

\subsection{Receptive-Filed Attention Convolution}

\begin{figure*}[ht]
	\centering
	\includegraphics[trim=0 0 0 0,clip,scale=0.43]{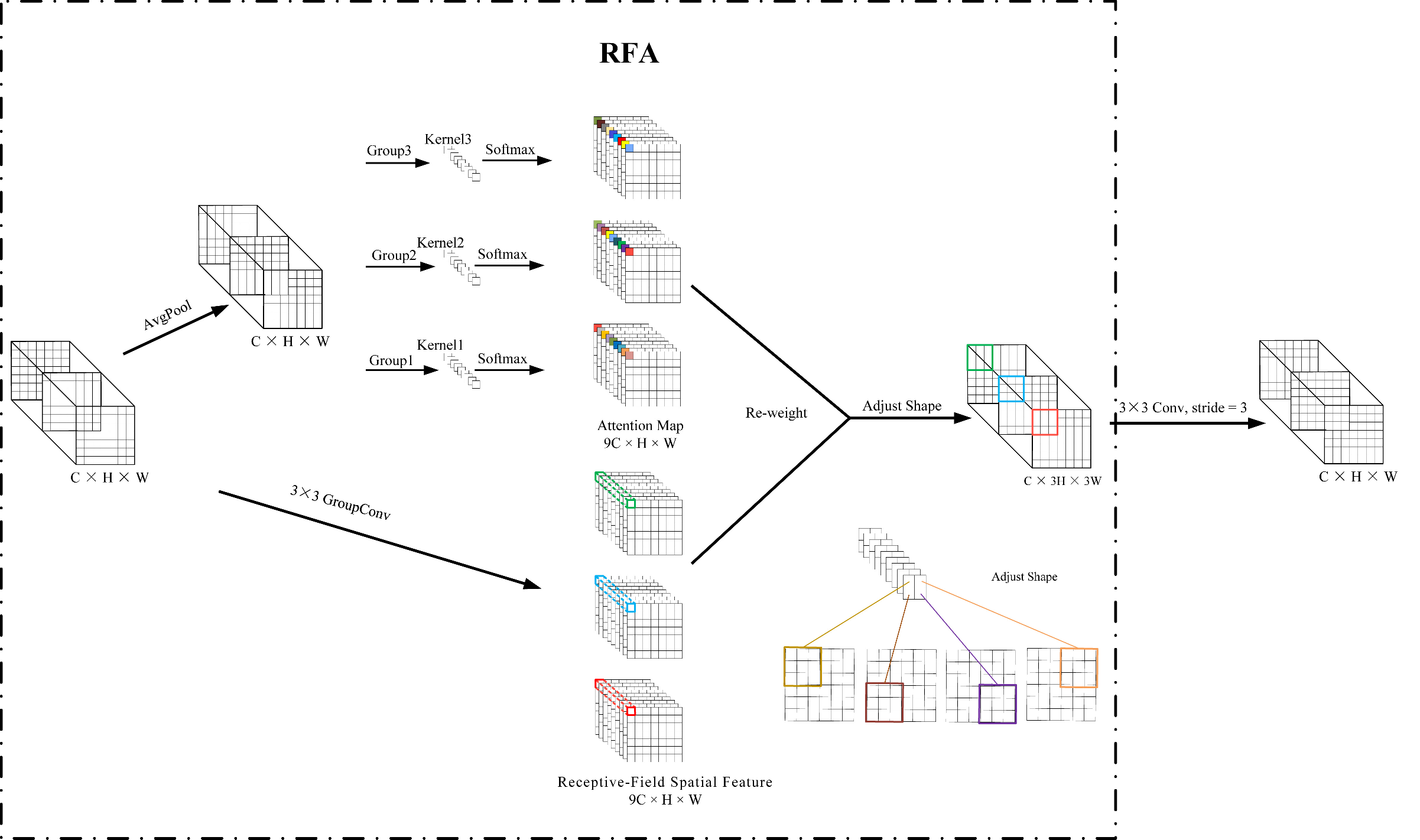}
	\caption{The detailed structure of RFAConv, which dynamically determines the importance of each feature in the receptive-field and solves the problem of parameters sharing. In the receptive-field spatial feature, the nine features adjacent to each channel represent the features in receptive-field slider at the original input corresponding position. Therefore, it needs to be adjusted to the spatial dimension by "Adjust shape" to facilitate the subsequent convolution operation.}
	\label{RFAConv}
\end{figure*}

Regarding the receptive-field spatial feature, we propose Receptive-Filed Attention (RFA) to address the limitations of the existing spatial attention mechanism and provides an innovative solution to the standard convolutional operation. Through this method, each receptive-field slider will receive a different attentional weight. Thus its combination with the standard convolution operation will transform the convolution operation into a non-shared parameter operator to improve the performance of the convolutional neural network. Inspired by the above, we propose Receptive-Field Attention Convolution (RFAConv). The overall structure of RFAConv with a 3 $\times$ 3 size convolutional kernel is shown in Figure~\ref{RFAConv} and C = 3 is an example.
Overall, the operational process of RFAConv can be divided into two steps, i.e., Grouping Optimization and Receptive-Field Attention Process. \\

\begin{figure}[h]
	\centering
	\includegraphics[trim=0 0 0 0,clip,scale=0.64]{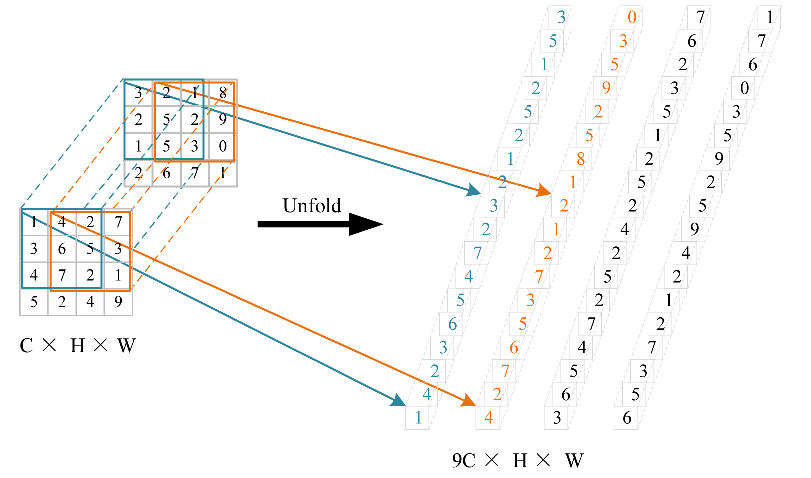}
	\caption{In the figure, it shows in detail an example of extracting 3 $\times$ 3 receptive-field spatial features by the Unfold method. It is worth noting that the unfold method performs the same zero-padding process on the feature map as the convolution, so the final output has the same width and height as the input. In the figure, we ignore the  zero-padding for the convenience of showing the process of the unfold method.}
	\label{unfold}
\end{figure}

\noindent \textbf{GroupConv Optimization:} As one of the popular frameworks in the field of deep learning, Pytorch provides the Unfold method  to extract receptive-field spatial features. The detailed structure is shown in Figure~\ref{unfold}, which extracts 3 $\times$ 3 receptive-field spatial features. Let input X $\in$ $ R^{C \times H \times W}$, after the Unfold method, its dimension becomes 9C $\times$ H$\times$W. Where C, H, and W represent the number of channels, height and width of input. The nine values adjacent to each other in the channel dimension are exactly the features in the receptive-field slider corresponding to the convolution operation in the original feature at each position. Although Unfold provided by Pytroch is able to extract receptive-field spatial features by parameterless way, it is slow. Therefore, in RFAConv, we utilize a fast method to extract receptive-field spatial features, i.e., GroupConv. As mentioned in the previous section, each 3 $\times$ 3 size window in the receptive-field spatial feature represents a receptive-field slider when the 3 $\times$ 3 convolution kernel is used to extract features, and similar to the unfold method in Figure~\ref{unfold}, the dimension of the output is 9C $\times$ H$\times$W. After using fast GroupConv to extract the receptive-field features become faster and more efficient than the original Unfold method.
As shown in Table~\ref{compare_rfaconv}, experiments based on the YOLOv5n and representative VisDrone dataset \cite{zhu2021detection} demonstrate it. It can be seen that RFAConv based on GroupConv obtains good performance while training 300 epochs requires less training time than the Unfold method. Moreover, we need to explain that the Unfold method is parameterless, while in the Table~\ref{compare_rfaconv}, it can be seen that the number of parameters required for the GroupConv-based method is the same as for the Unfold method. Because we use a lightweight approach to interacting information in receptive-field to reduce parameters.

\begin{table*}[h]
	\centering
	\small
	\caption{Object detection experiments based on YOLOv5n and VisDrone datasets to illustrate the advantages of RFAConv built on GroupConv.}
	\renewcommand\arraystretch{0.5}
	\setlength{\tabcolsep}{1mm}{
		\begin{tabular}{lccccc}
			\toprule
			Methods                          & mAP50($\%$) & mAP($\%$) & FLOPS(G) & Param(M)  &Training Time (Hours) \\
			\midrule
			YOLOv5n                            &26.43  & 13.66   & 4.3   &1.78 & 6.81 \\
			YOLOv5n + RFAConv (Unfold)         &27.43  & 14.22 & 4.6   &1.85  & 10.42 \\
			YOLOv5n + RFAConv (GroupConv)      &27.58  & 14.36 & 4.7   &1.85  & 7.37 \\
			\bottomrule
	\end{tabular}}
	\label{compare_rfaconv}
\end{table*}

\begin{figure}[h]
	\centering
	\includegraphics[trim=0 0 0 0,clip,scale=1.9]{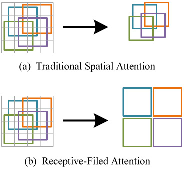}
	\caption{(a) and (b) denote the attentional weights obtained by traditional spatial attention and receptive-field attention, respectively. It is evident that traditional spatial attention does not unfold the receptive-slider features, which leads to problems with sharing the obtained attentional weights.}
	\label{attention-comapre}
\end{figure}

\noindent \textbf{Receptive-Field Attention Process:} Recent research has shown that interacting information can enhance network performance, as demonstrated in \cite{10891570}. Similarly, for RFAConv, interacting receptive-field feature information to learn the attention map can enhance network performance. However, interacting with each receptive-field feature can result in additional computational overhead, so to minimize computational overhead and number of parameters, the AvgPool is utilized to aggregate the global information of each receptive-field feature. Then, a 1 $\times$ 1 group convolutional operation is used to interact information. Finally, we use softmax to emphasize the significance of each feature within the receptive-field feature. The range of features targeted by softmax are the features of the corresponding receptive-field slider on each channel, which is analogous to grouping features by the number of channels. In general, the calculation of RFA can be expressed as:
\begin{equation}
\small
\begin{aligned}
&F = Softmax(g^{1\times 1}(AvgPool(X))) \times  ReLU(Norm(g^{k\times k}(X)))\\
&~~~=  A_{rf} \times  F_{rf}\\
\end{aligned}
\end{equation}
Here, $g^{i \times i}$ represents a grouping convolution of size $i\times i$, k represents the size of the convolution kernel, Norm stands for normalization, $X$ represents the input feature maps, $F$ is obtained by multiplying the attention map $A_{rf}$ with the transformed receptive-field spatial feature $F_{rf}$. Unlike traditional spatial attention, such as  CBAM and CA, RFA assigns different weights to each feature in the receptive-field slider by softmax operation, and the weights obtained by learning are not shared in each convolutional slider after extension, as shown in Figure~\ref{attention-comapre}. Attentional weights learned by traditional spatial attention mechanisms are shared across each receptive-field slider. In contrast, the attentional weights learned by the RFA do not overlap, so each receptive-field slider has independent attentional weights that do not share weights from other sliders. The performance of convolutional neural networks is limited by the standard convolutional operations due to the fact that the convolution operation relies on shared parameters and is not sensitive to differences in information brought about by positional variations. However, RFAConv can completely address this issue by emphasizing the significance of different features within the receptive-field slider and prioritizing the receptive-field spatial feature. The feature map obtained through RFA is the receptive-field spatial feature, which does not overlap after "Adjust Shape" (It transforms the receptive-field slider stacked on the channel dimension into the spatial dimension to complete the subsequent convolution operation). 

After the two-step process described above, the features are k times in height and width, requiring a k $\times$ k convolution operation with a stride = k to extract feature information. The process of extracting features in the receptive-field spatial feature by a k $\times$ k convolution operation with a stride = k just corresponds to the process of convolutional operation in original spatial features. RFA focuses on the receptive-field spatial feature, so the weight of the attention gained is not shared among each receptive-field slider. Therefore, when we consider the last convolution operation as a whole with the previous attention weighting operation, the problem of convolution parameter sharing is perfectly solved. Be careful not to look at the subsequent convolution operation separately, but to see the attention weighting operation and the convolution operation as a whole. The convolution operation RFAConv designed by RFA brings good gains for convolution, and it innovates the standard convolution.

Furthermore, We assert that existing spatial attention mechanisms, such as CBAM \cite{woo2018cbam} and CA  \cite{hou2021coordinate}, should prioritize receptive-field spatial features to improve network performance. As is well-known, the network model based on the self-attention mechanism \cite{TIAN2025129148} has achieved great success, because it solves the problem of convolution parameter sharing and models long-distance information. However, the self-attention mechanism also introduces significant computational overhead and complexity to the model. We argue that directing the attention of existing spatial attention mechanisms to the receptive-field spatial feature can solve the problems of parameter sharing and modeling of long-range information in a way similar to self-attention. This approach requires significantly fewer parameters and computational resources than self-attention. The answers are as follows: 

(1) The combination of the spatial attention mechanism, which focuses on the receptive-field spatial feature, with convolution eliminates the problem of convolution parameter sharing. 
(2) The current spatial attention mechanism already considers long-distance information and can obtain global information through global average pooling or global maximum pooling, which explicitly takes into account long-range information. 

\begin{figure*}[h]
	\centering
	\includegraphics[trim=0 0 0 0,clip,scale=0.67]{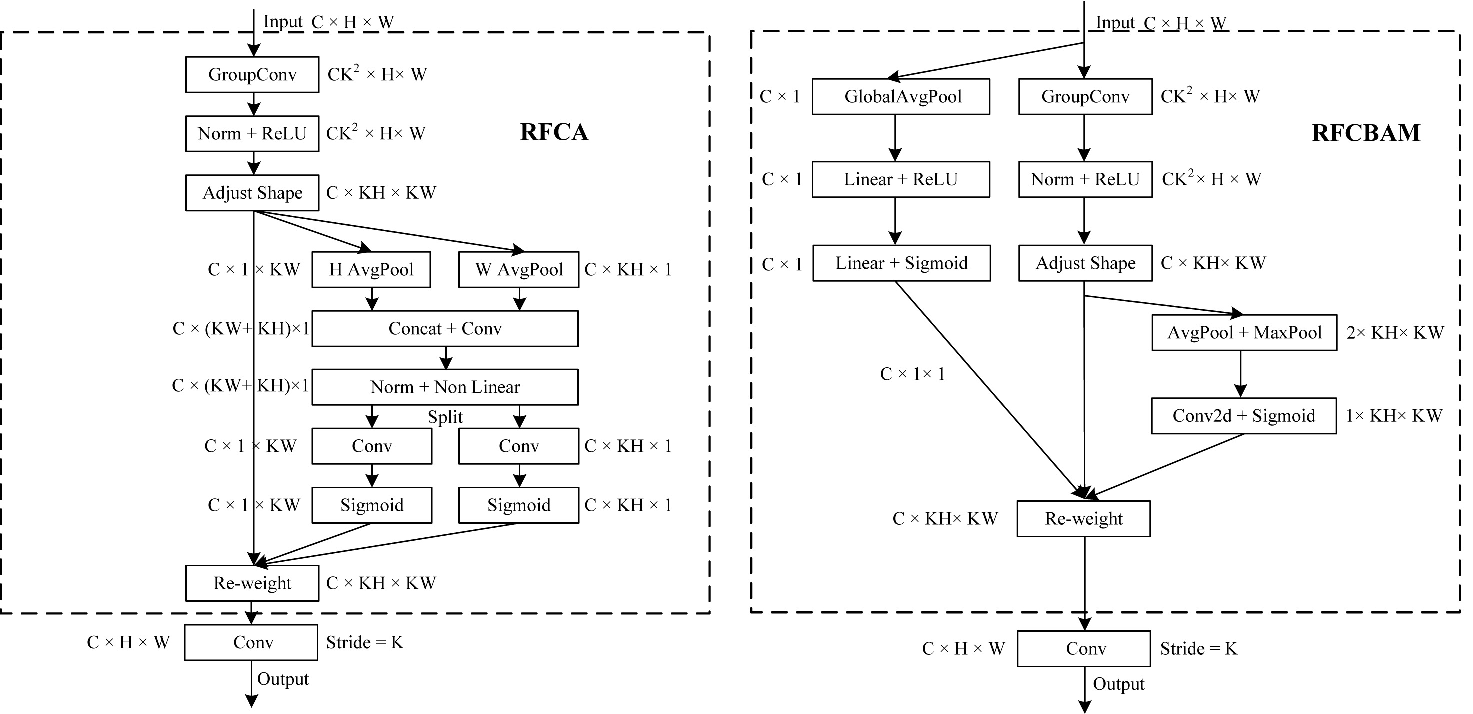}
	\caption{Detailed structure of RFCAConv and RFCBAMConv, which focus on receptive-field spatial features. Comparing the original CBAM, We use SE attention to replace CAM in RFCBAM. Because, this can reduce computational overhead.}
	\label{RFCAConv-RFACBAMConv}
\end{figure*}

Therefore, we design new CBAM and CA models called RFCBAM and RFCA, which focus on the receptive-field spatial feature. Similar to RFA, the final convolution operation of $k \times k$ with stride = k is used to extract the feature information. The specific structure of these two new convolution methods, as shown in Figure~\ref{RFCAConv-RFACBAMConv}, we call these two new convolution operations RFCBAMConv and RFCAConv. Comparing the original CBAM, We use Squeeze-and-Excitation attention to replace CAM in RFCBAM. Because this can reduce computational overhead. Moreover in RFCBAM, channel and spatial attention are not performed in separate steps. Instead, they are weighted simultaneously, allowing the attention map obtained for each channel to be different.

\section{Experiments and Discussions}
To verify the effectiveness of our method, we perform classification, object detection and semantic segmentation experiments. The equipment for all experiments are based on RTX3090. In the classification experiments on ImageNet-1k, we use four RTX3090 to train the model in parallel. In the experiments, Cuda version and Pytroch version are 11.1 and 1.8.1, respectively.

\subsection{Classification experiments on ImageNet-1k}
Currently, for classification tasks, ImageNet-1k is a large-scale dataset that contains 1.28 million training sets, and 50,000 validation sets. Therefore, We perform experiments to validate our method on the ImageNet-1k. Similar to RFAConv, we construct CBAMConv and CAConv by combining the CBAM and CA, respectively, with an additional 3 $\times$ 3 convolution layer at the end of the CBAM and CA modules. We also compare the CAMConv, which is constructed using the channel attention mechanism CAM \cite{woo2018cbam}. we evaluate in ResNet18, ResNet34. Specifically, RFAConv, CBAMConv, CAConv, and CAMConv are used to replace (r) the first convolutional layer of BasicBlock in ResNet18 and ResNet34, respectively. In general, the new convolution is structured as shown in Table~\ref{resnet}.
\begin{table*}[h]
	\centering
	\small
	\caption{The ResNet18 and ResNet34 are construct by the new convolution operation.}
	\setlength{\tabcolsep}{4mm}{
		\begin{tabular}{lccc}
			\toprule
			Layer Name & Output Size & ResNet18 & ResNet34    \\
			\midrule
			Conv1    & 112 $\times$ 112 &  & \\
			\midrule
			Layer1    & 56 $\times$ 56 &  $\begin{bmatrix}NewConv & 3 \times 3\\ Conv & 3 \times 3 \end{bmatrix} \times 2$ & $\begin{bmatrix}NewConv & 3 \times 3\\ Conv & 3 \times 3 \end{bmatrix} \times 3$\\
			\midrule
			Layer2    & 28 $\times$ 28 &  $\begin{bmatrix}NewConv & 3 \times 3\\ Conv & 3 \times 3 \end{bmatrix} \times 2$ & $\begin{bmatrix}NewConv & 3 \times 3\\ Conv & 3 \times 3 \end{bmatrix} \times 4$\\
			\midrule
			Layer3    & 14 $\times$ 14 &  $\begin{bmatrix}NewConv & 3 \times 3\\ Conv & 3 \times 3 \end{bmatrix} \times 2$ & $\begin{bmatrix}NewConv & 3 \times 3\\ Conv & 3 \times 3 \end{bmatrix} \times 6$\\
			\midrule
			Layer4    & 7 $\times$ 7 &  $\begin{bmatrix}NewConv & 3 \times 3\\ Conv & 3 \times 3 \end{bmatrix} \times 2$ & $\begin{bmatrix}NewConv & 3 \times 3\\ Conv & 3 \times 3 \end{bmatrix} \times 3$\\
			\midrule
			& 1 $\times$ 1 & \multicolumn{2}{c}{ AvgPool 1000-d }\\	
			\bottomrule
	\end{tabular}}
	\label{resnet}
\end{table*}

The NewConv in Table~\ref{resnet} represents the convolution mode constructed by the attention mechanism. For CAConv, CBAMConv, and CAMConv, it can be considered that CA, CBAM, and CAM are added before the first layer of convolution in BasicBlock. In the image classification experiments, we train 100 epochs for each model with  batch-size of 128. The learning rate start from 0.1 and decrease every 30 epochs, 0.1 times each time. In experiments, we follow most of the previous work and report the accuracy for TOP1 and TOP5, respectively. Table~\ref{classification1} shows the results produced by different networks on the ImageNet-1K validation set. It is clear that replacing the 3 $\times$ 3 convolutional operation with RFAConv significantly improves the recognition results. Compared to the baseline models RestNet18 and ResNet34 , the network constructed by RFAConv achieves the best recognition results at the cost of only a small increase in parameters and computational overhead. Such as ResNet18 constructed based on RFAConv only adds 0.16 M parameters and 0.09 G computational overhead over the original model, and increases the accuracy by 1.64\% and 1.24\% on TOP1 and TOP5, respectively. Whether it is RFAConv, CBAMConv, CAConv, or other convolutional operations constructed for attention mechanisms, it can be seen as adding attention mechanisms in front of convolution. So to compare the work of other attention mechanisms added to ResNet, we also compare ECA \cite{wang2020eca}, QEA \cite{zhang2023qea}, QCA \cite{zhang2023qca}, EMAS \cite{sheng2023efficient}, TA \cite{Misra_2021_WACV} and GAM \cite{liu2021global} in Table~\ref{classification1}. Compared to these works, RFAConv can still get good performance.

\begin{table}[H]
	\centering
	\small
	\footnotesize
	\caption{Classification results on ImageNet-1K using the ResNet18 and ResNet34. The different convolutional operation constructed by the attention mechanism is compared. Moreover, some related works are compared. }
	\renewcommand\arraystretch{0.5}
	\setlength{\tabcolsep}{1mm}{
		\begin{tabular}{lcccc}
			\toprule
			Models & FLOPS(G) & Param(M) & Top1(\%) & Top5(\%)    \\
			\midrule
			ResNet18          & 1.82   & 11.69      & 69.59    &89.05  \\
			+ CAMConv(r)     & 1.83   & 11.75      &70.76     &89.74  \\
			+ CBAMConv(r)    & 1.83   & 11.75      &69.38     &89.12  \\
			+ CAConv(r)      & 1.83   & 11.74      &70.58     &89.59  \\
			+ GAM		 	 & 2.45   & 16.04      &70.66     &89.77  \\
			+ QEA            & 1.70   & 11.17      &70.55     &89.92  \\
			+ QCA            & 1.70   & 11.19      &70.84     &89.92  \\
			+ EMAS         	 & 1.81   & 11.69      &71.04     &90.02  \\
			+ TA		 	 & 1.83   & 11.69      &71.09     &89.99  \\
			+ RFAConv(r)     & 1.91   & 11.85      &\textbf{71.23}     &\textbf{90.29}  \\
			
			\midrule
			ResNet34         & 3.68   & 21.80      &73.33     &91.37  \\
			+ CAMConv(r)     & 3.68   & 21.93      &74.03     &91.69  \\
			+ CBAMConv(r)    & 3.68   & 21.93      &72.95     &91.26  \\
			+ CAConv(r)      & 3.68   & 21.91      &73.76     &91.68  \\
			+ ECA			 & 3.68   & 21.80      &74.21     &91.88  \\	
			+ QEA            & 3.43   & 20.82      &73.76     &91.52  \\
			+ EMAS         	 & 3.66   & 21.80      &74.14    &91.88   \\
			+ RFAConv(r)     & 3.84   & 22.16      &\textbf{74.25}     &\textbf{92.03}  \\
			\bottomrule
	\end{tabular}}
	\label{classification1}
\end{table}

Moreover, as we mentioned earlier, spatial attention can be enhanced again by focusing on receptive-field spatial features. Therefore, we design RFCBAMConv and RFCAConv, which are improvements of CBAM and CA. In order to verify their advantages, we conduct experiments based on ResNet18 and report the relevant data in Table~\ref{classification2}. It is obvious that RFCBAMConv and RFCAConv achieve better recognition accuracy, compared with CBAMConv and CAConv in Table~\ref{classification1}. Most importantly, they also significantly improve performance at the cost of only a small increase in parameters and computational overhead. This is a strong demonstration that spatial attention can be improved by placing attention into the receptive-field spatial features. It fully demonstrates that spatial attention can be improved by placing attention into the receptive-field spatial features. In the Table~\ref{classification2}, we have also compared other excellent works. Compared to CBAM, works such as AFF-CAM \cite{Yang_2022_ACCV}, MCA \cite{yu2023mca}, LMA \cite{yu2023lma} and DAS \cite{salajegheh2023deformable} have been able to obtain satisfactory results for the network. But the performance obtained from RFCBAM obtained by focusing on receptive-field spatial features to improve CBAM and combining convolutional operations has obtained better performance.

\begin{table}[h]
	\centering
	\footnotesize
	\caption{RFCBAMConv and RFCAConv improve the performance of CBAMConv and CAConv, and significantly improve the accuracy obtained by the network compared to other excellent works. The table shows that the classification accuracy is significantly improved on ImageNet-1k.}
	\renewcommand\arraystretch{0.5}
	\setlength{\tabcolsep}{1mm}{
		\begin{tabular}{lcccc}
			\toprule
			Models & FLOPS(G) & Param(M) & Top1(\%) & Top5(\%)    \\
			\midrule
			ResNet18         & 1.82   & 11.69      & 69.59    &89.05  \\
			+ AFF-CAM        & 1.84   & 11.84      & 71.52    & 90.36 \\
			+ MCA            & 1.82   & 11.69      & 71.57    & 90.39 \\
			+ LMA            & 1.82   & 11.69      & 71.43    & 90.18 \\
			+ DAS            & 1.826   & 11.82     & 72.03    & 90.70 \\
			+ RFCAConv(r)    & 1.92   & 11.89      &72.01     &90.64 \\
			+ RFCBAMConv(r)  & 1.90   & 11.88      &\textbf{72.15}    &\textbf{90.71}  \\
			\bottomrule
	\end{tabular}}
	\label{classification2}
\end{table}
All classification experiments clearly demonstrate the significant advantages of our methods, because RFAConv, RFCBAMConv, and RFCAConv completely address the problem of convolution kernel parameter sharing. Moreover, it is worth noting that RFCBAMConv and RFCAConv outperform RFAConv, because they not only solve the problem of convolution kernel parameter sharing but also consider long-distance information through global pooling.

\begin{figure}[h]
	\centering
	\includegraphics[trim=0 0 0 0,clip,scale=0.9]{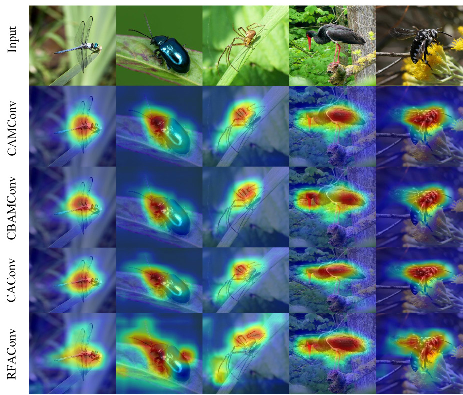}
	\caption{Each network is built on ResNet18 based on attention convolution, and the construction process is shown in Table~\ref{resnet}. We use Grad-CAM as our visualization tool to visualize networks without the last layer of classifiers. Compared to other attention convolution methods, our RFAConv can help the network to better recognize and highlight the key regions of objects}
	\label{Grad-CAM1}
\end{figure}

Moreover, to provide a more intuitive analysis, as with the most work, we use the Grad-CAM \cite{selvaraju2017grad} algorithm for visualization. Grad-CAM highlights the regions of interest of different networks for a particular class of objects. To some extent, it is possible to see how the networks utilize the features. We randomly selected some images in the validation set of ImageNet-1K and visualized the results of the networks constructed with different attention convolutions based on ResNet18 separately. As shown in the Figure~\ref{Grad-CAM1}, compared with other attention convolutions, our RFAConv can help the network to better recognize and highlight the key regions of objects.

\begin{figure}[h!]
	\centering
	\includegraphics[trim=0 0 0 0,clip,scale=0.95]{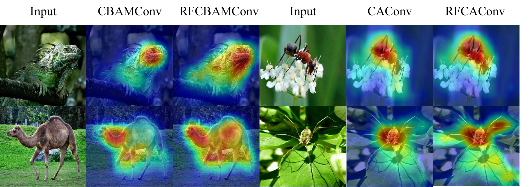}
	\caption{We put the attention of CBAM and CA into the receptive-field spatial features and improve them to obtain RFCBAM and RFCA. Then RFCBAMConv and RFCAConv are constructed by the same method as RFA. Obviously, compared to CBAMConv and CAConv, the improved obtained RFCBAMConv and CAConv can help the network to better recognize and highlight the key regions of objects.}
	\label{Grad-CAM2}
\end{figure}

\begin{figure}[h!]
	\centering
	\begin{subfigure}
		\centering
		\includegraphics[width=0.45\linewidth]{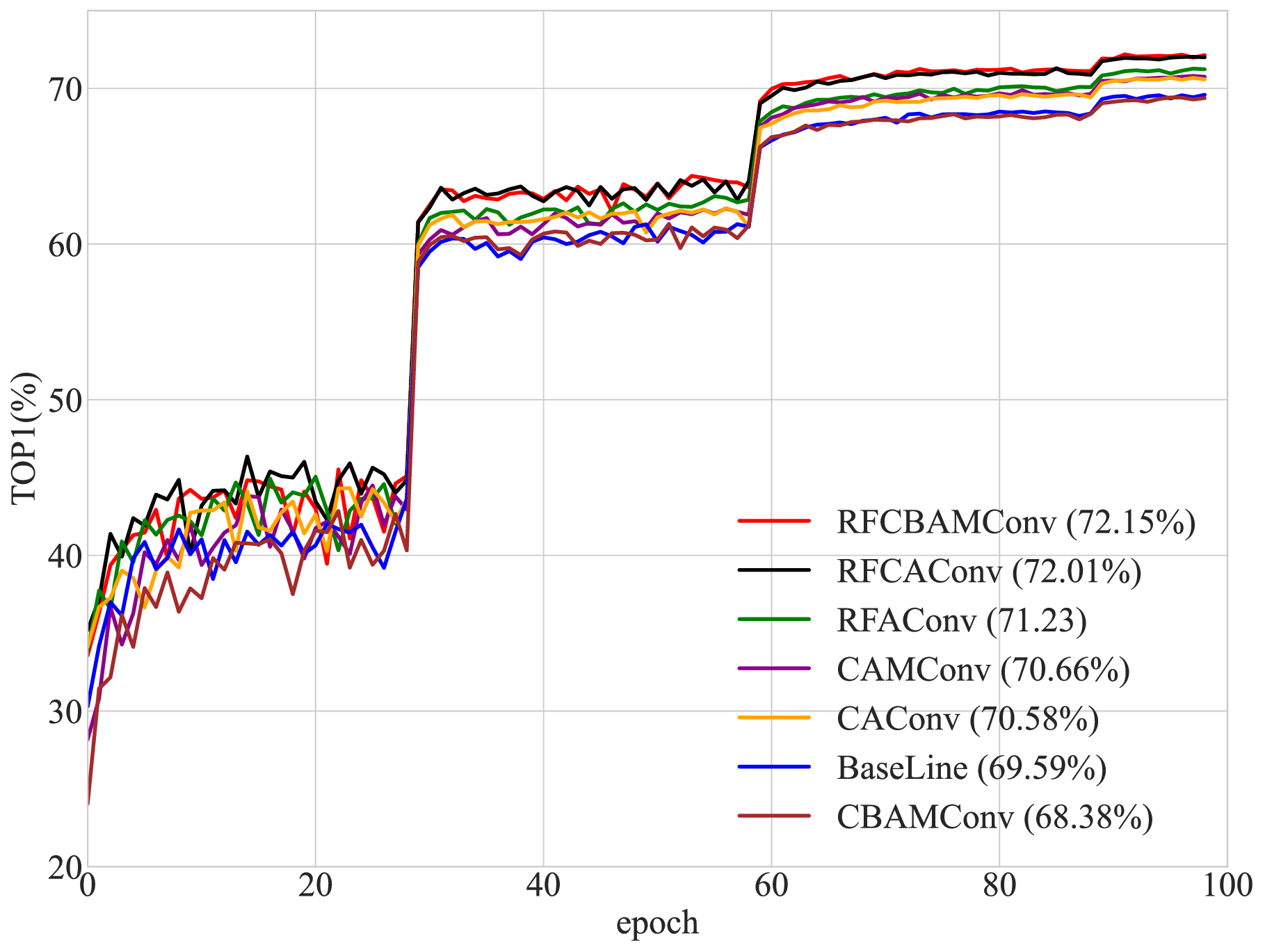}
	\end{subfigure}
	\centering
	\begin{subfigure}
		\centering
		\includegraphics[width=0.45\linewidth]{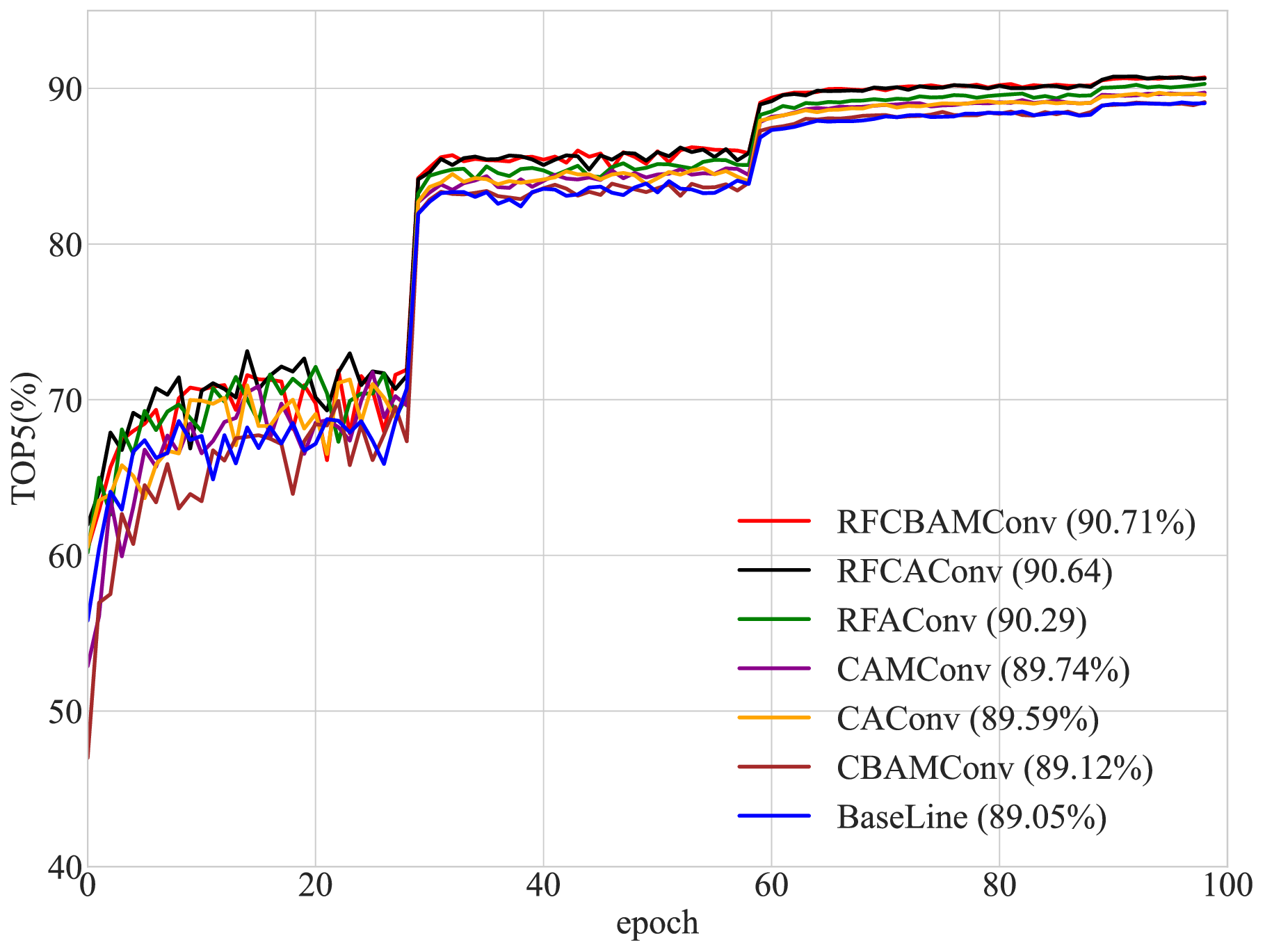}
	\end{subfigure}
	\caption{It shows the accuracy change curves of ResNet18 for TOP1 and TOP5 on the validation set during the training process.}
	\label{top-changes}
\end{figure}

We also use Grad-CAM to visualize ResNet18 constructed by CBAMConv, RFCBAMConv, CAConv, and RFCAConv. RFCBAM is obtained by improving the attention of CBAM by putting it into the receptive-field spatial feature. Similarly, RFCA is obtained by CA after the same method. As shown in Figure~\ref{Grad-CAM2}, the improved RFCA and RFCBAM  can help the network to better recognize and highlight the key regions of objects after combining with the convolution operation. Moreover, to further observe the accuracy changes during the training process of networks constructed by different attention convolutions, we visualized the TOP1 and TOP5 data obtained by ResNet18 on the validation set. As shown in Figure~\ref{top-changes}, it is evident that compared to other methods, convolutional operations focusing on receptive-field spatial features can consistently maintain a higher level of accuracy in the network.

\subsection{Classification experiments on ImageNet-200 and Places365}
In order to verify that the proposed method can improve the performance of different networks, we conduct corresponding experiments based on the ImageNet-200 dataset. Compared with ImageNet-1k, ImageNet-200 contains 100000 training sets and 10000 validation sets, which can again verify whether the proposed method can gain advantages even with small amount of data. In experiments, we select six networks of different sizes for the corresponding classification experiments, such as FasterNet \cite{chen2023run}, InceptionNeXt \cite{yu2024inceptionnext}, MobileNet \cite{howard2019searching}, ShuffleNet \cite{ma2018shufflenet}. For these networks, we utilize the proposed convolution operation to replace the large-size standard convolution operation in their networks for comparison experiments. In the experiment, all the parameter settings are similar to the previous section. Table~\ref{classfication200} shows the corresponding experimental results. Compared with the baseline model, after replacing the standard convolution operation with the proposed convolution operation, the classification performance is better than the baseline model, despite the increase in the number of small parameters and computational overhead. Therefore, whether it is the large-scale dataset ImageNet-1K or the smaller dataset ImageNet-200, our proposed methods are able to obtain good performance. The reason is that different images exhibit varying information disparities at different positions, yet the parameter-sharing mechanism of standard convolution fails to capture these distinct information patterns. In contrast, the proposed convolution operation can employ distinct convolution kernels at different spatial positions, thereby adaptively capturing these location-specific information disparities.

\begin{table*}[H]
	\centering
	\footnotesize
	\caption{Classification results on ImageNet-200 using different baseline model. Different networks constructed by the  RFCAConv are compared.}
	\renewcommand\arraystretch{0.5}
	\setlength{\tabcolsep}{1mm}{
		\begin{tabular}{lcccc}
			\toprule
			Models & FLOPS(G) & Param(M) & Top1(\%) & Top5(\%)    \\
			\midrule
			FasterNet		 & 157.53   & 13.96    & 41.28    & 67.88  \\
			FasterNet + RFCAConv(r)    & 161.58   & 14.12    & 41.84    & 68.17  \\
			InceptionNeXt    & 685.42   & 47.52    & 20.26    & 45.01  \\
			InceptionNeXt + RFCAConv		 & 690.98   & 47.73    & 28.03    & 55.00  \\
			MobileNet      & 2.79     & 0.61     & 36.26    & 63.56  \\
			MobileNet + RFCAConv(r)	 & 5.48     & 0.72     & 40.28    & 67.78  \\
			ShuffleNet       & 12.22    & 1.11     & 42.23    & 68.66  \\
			ShuffleNet + RFCAConv(r)	 & 18.11    & 1.38     & 43.68    & 69.64  \\

			\bottomrule
	\end{tabular}}
	\label{classfication200}
\end{table*}

Moreover, to further compare other pure attention-based architectures and hybrid architectures, we construct a corresponding network based on InceptionNeXt by replacing some large-sized convolutions with RFCAConv, and conduct corresponding experiments on the large-scale scene recognition dataset Places365. Places365 contains a total of 365 categories, with over 1.8 million images in total. Table~\ref{classfication365} presents the corresponding experimental results, which are basically consistent with the results in \cite{yu2025mambaout}. The results obtained by InceptionNeXt-A \cite{yu2024inceptionnext} are similar to those achieved by the pure attention mechanism architecture. However, in the Places365 dataset, the classification accuracy of InceptionNeXt-A is lower compared to other networks. This is understandable, as Places365 is a scene classification dataset that requires more contextual information. Therefore, compared to other attention mechanism networks, InceptionNeXt-A achieves lower accuracy. However, after incorporating RFCAConv into InceptionNeXt-A, our network achieves better results than PVTv2-B0 \cite{wang2022pvt},  DeiT-T \cite{touvron2021training}, and T2T-ViT-7 \cite{yuan2021tokens} under nearly identical parameters and computational overheads.

\begin{table}[H]
	\centering
	\small
	\footnotesize
	\caption{Results of classification using different network architectures on the Places365 dataset.}
	\renewcommand\arraystretch{0.5}
	\setlength{\tabcolsep}{1mm}{
		\begin{tabular}{lcccc}
			\toprule
			Models & FLOPS(G) & Param(M) & Top1(\%) & Top5(\%)    \\
			\midrule
			PVTv2-B0        & 0.53   &  3.5     &  34.25   & 64.28 \\
			DeiT-T          & 1.07   &  5.56    &  36.33   & 66.72 \\
			T2T-ViT-7       & 0.98   &  4.09    &  37.56   & 68.75 \\
			InceptionNeXt-A & 0.5    &  3.54    &  35.67   & 65.78 \\
			Ours            & 0.52   &  3.6     &  \textbf{38.14}   & \textbf{69.1} \\
			\bottomrule
	\end{tabular}}
	\label{classfication365}
\end{table}

\subsection{Object detection experiments on COCO2017}
We conduct object detection experiments on COCO2017 to re-evaluate our methods. COCO2017 contains 118287 training sets and 5000 verification sets. We select YOLOv5n, YOLOv7-tiny, and YOLOv8n models to perform a series of experiments.

All parameters except for epoch and batch-size are set to the default values. We train each model for 300 epochs with a batch-size of 32. To be similar in classification, we replace some convolution operation in the baseline model with novel convolution operations constructed using attention mechanisms, such as CAM, CBAM, and CA. Moreover, to fairly compare the performance of TA \cite{Misra_2021_WACV}, HAM \cite{li2022ham}, MCA \cite{yu2023mca}, the same convolutional operations are added behind these attention mechanisms to constitute TAConv, MCAConv and HAMConv. Similarly, we also use them to replace the convolutional operations at the same locations to conduct related experiments. Specifically, we replace all 3 $\times$ 3 convolution operations in the yaml files for YOLOv5 and YOLOv8 using attention convolution. And for YOLOv7, we replace the first 3 $\times$ 3 convolution operation in all ELAN \cite{wang2023yolov7} in the backbone. Note that HAMConv replacing the convolution operation at the same location in YOLOv8n would create a NaN problem in training, so we do not use HAMConv in YOLOv8n.

\begin{figure*}[h!]
	\centering
	\begin{subfigure}
		\centering
		\includegraphics[width=0.325\linewidth]{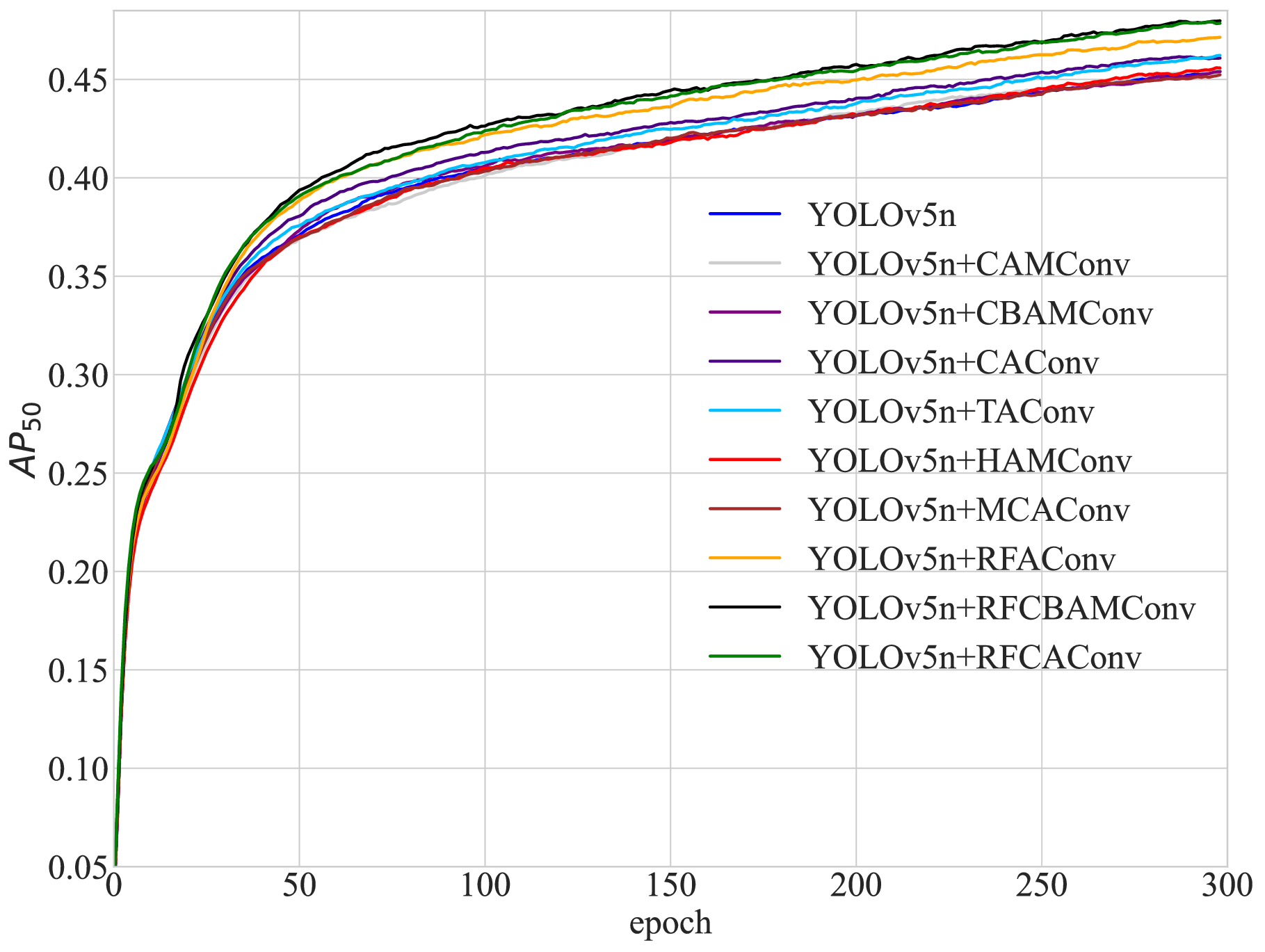}
	\end{subfigure}
	\centering
	\begin{subfigure}
		\centering
		\includegraphics[width=0.325\linewidth]{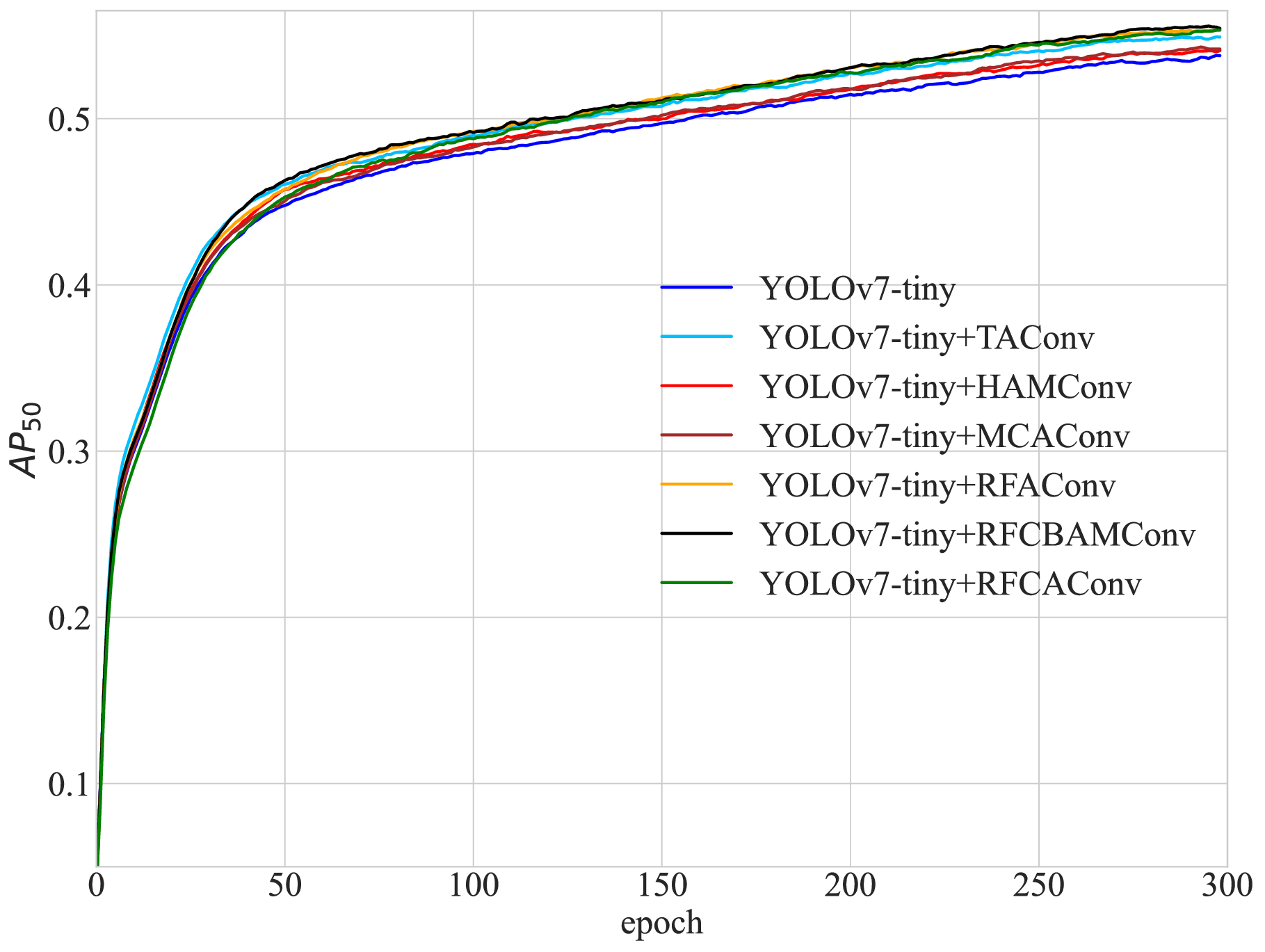}
	\end{subfigure}
	\centering
	\begin{subfigure}
		\centering
		\includegraphics[width=0.325\linewidth]{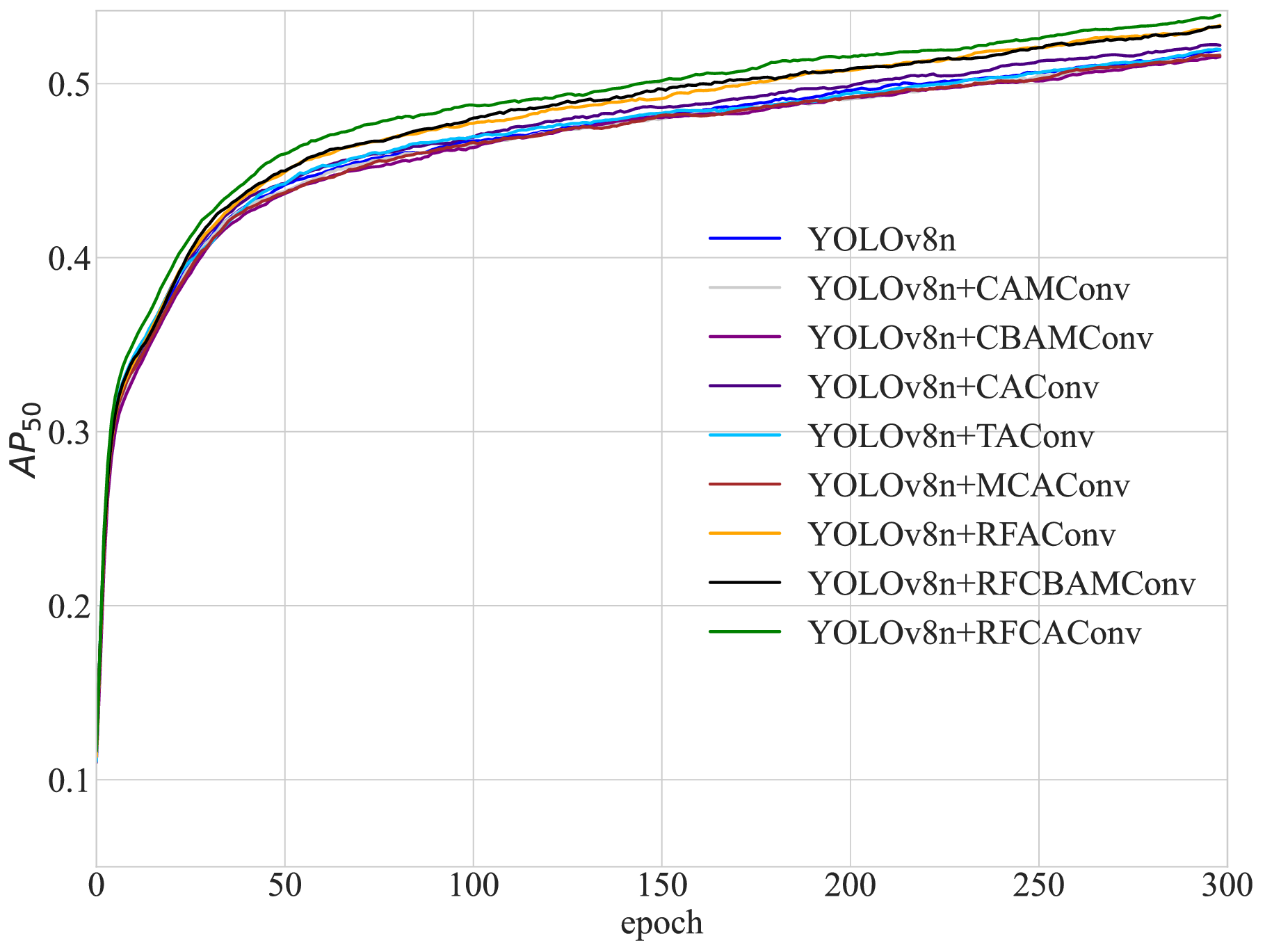}
	\end{subfigure}
	\centering
	\begin{subfigure}
		\centering
		\includegraphics[width=0.325\linewidth]{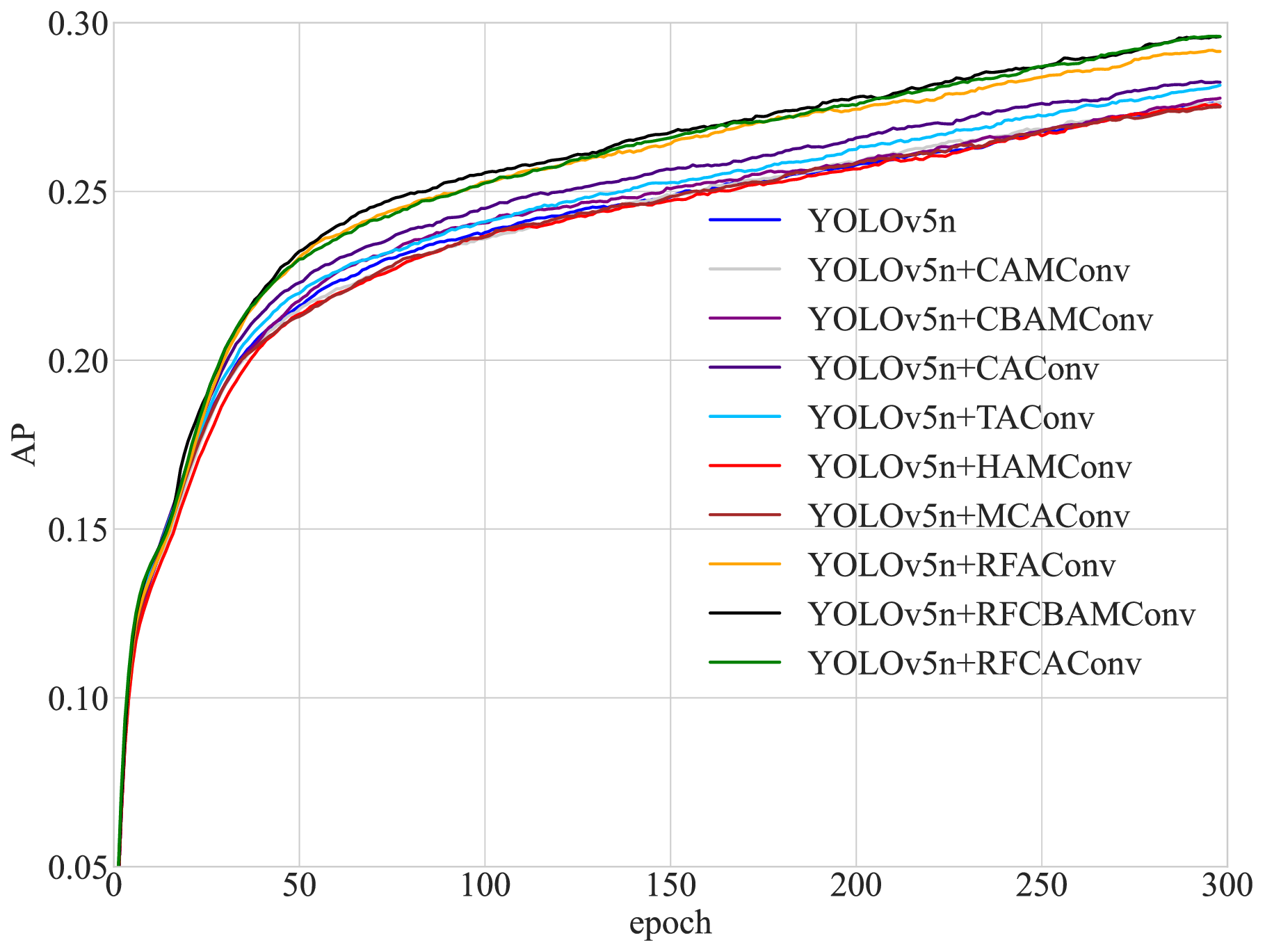}
	\end{subfigure}
	\centering
	\begin{subfigure}
		\centering
		\includegraphics[width=0.325\linewidth]{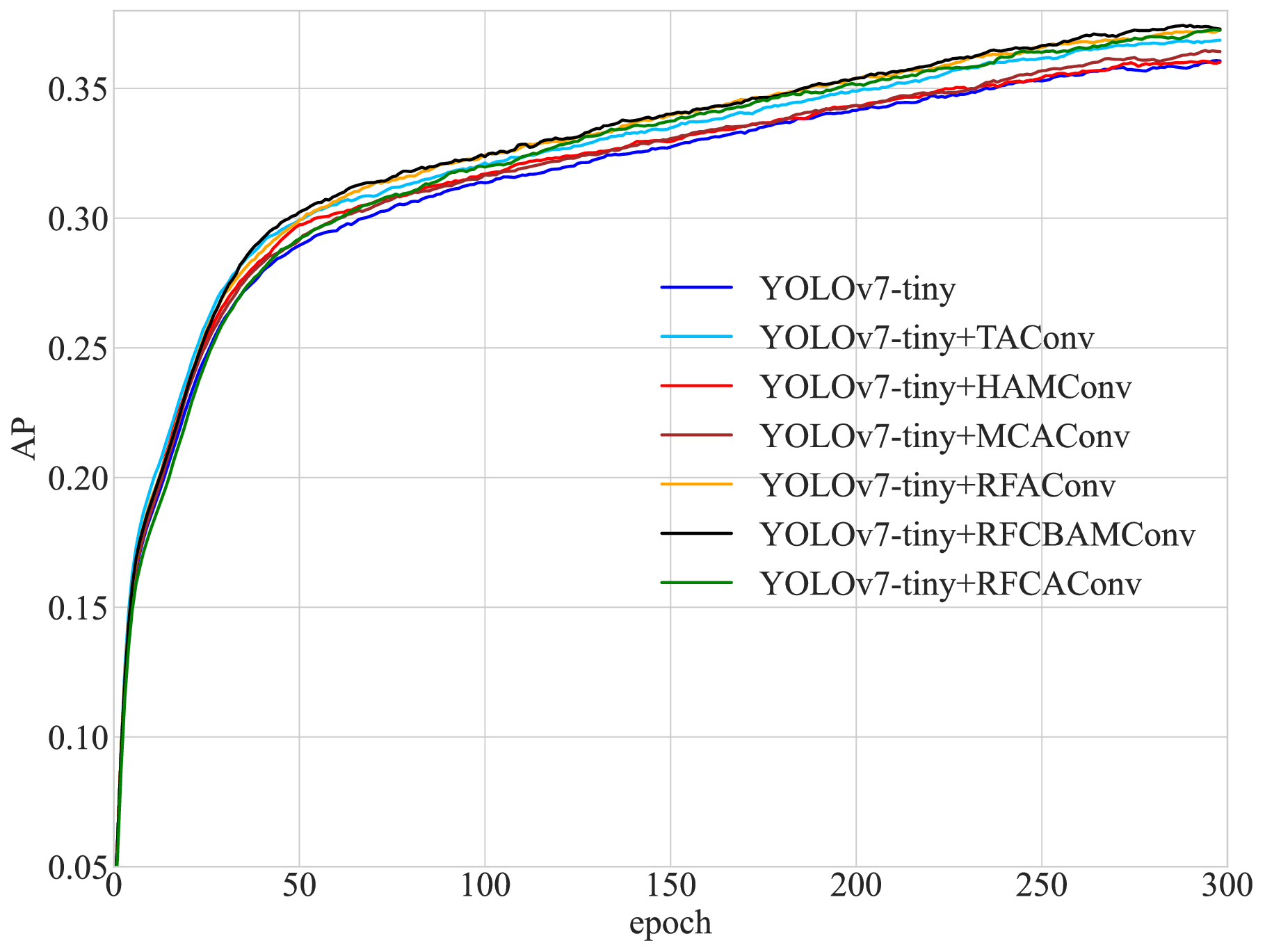}
	\end{subfigure}
	\centering
	\begin{subfigure}
		\centering
		\includegraphics[width=0.325\linewidth]{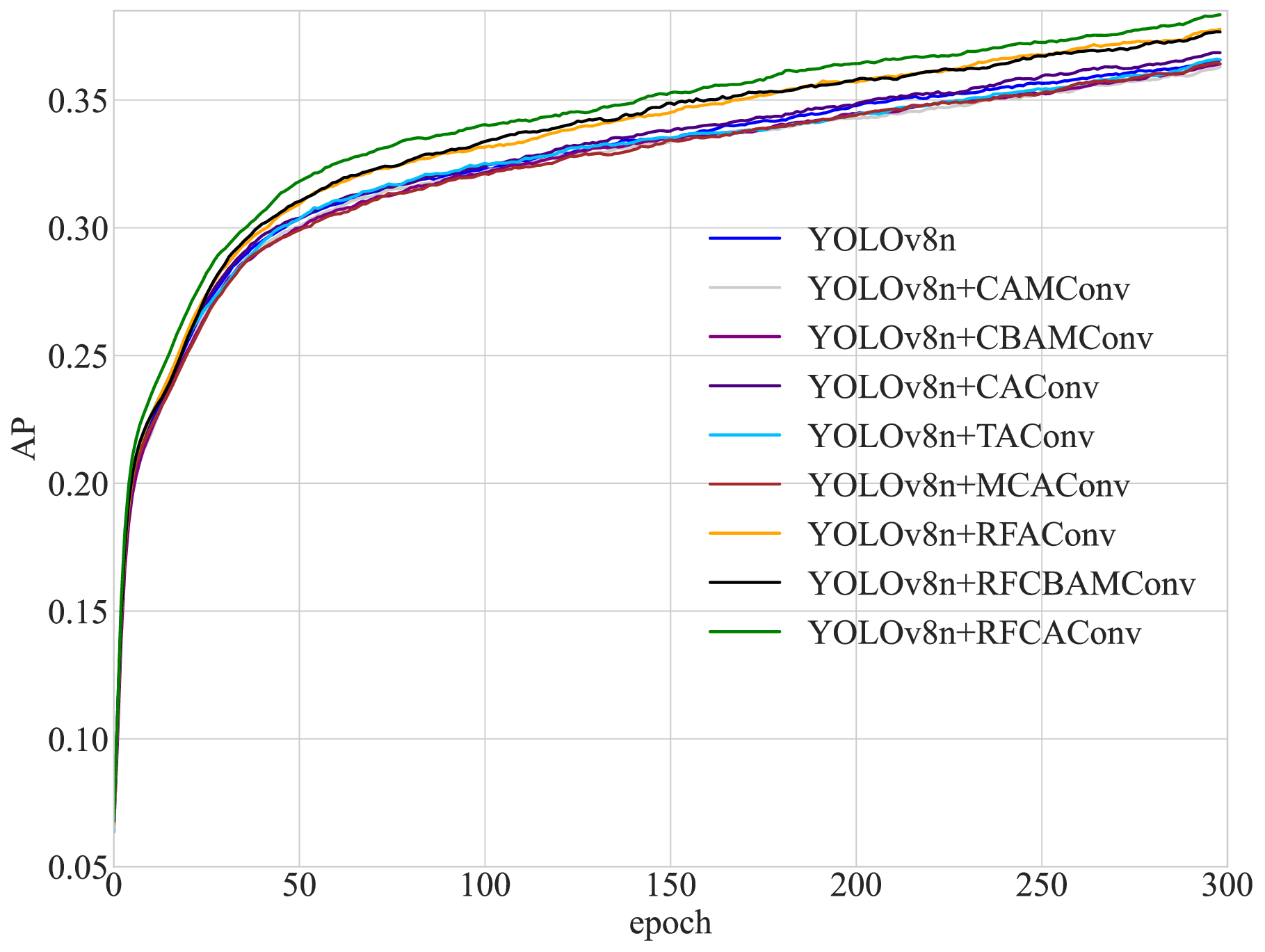}
	\end{subfigure}
	\caption{The $AP_{50}$ and $AP$  change during training for the different YOLOv5n, YOLOv7-tiny and YOLOv8n constructed by attention convolution.}
	\label{AP_change}
\end{figure*}

Following the previous work, we report $AP_{50}$, $AP_{75}$, $AP$, $AP_{S}$, $AP_{M}$ and $AP_{L}$ separately. Moreover, in order to better display the performance of different networks, we chose the training process of detection network for visualization. We visualize how the AP50 and AP changes with the number of iterations. The experimental results are shown in Table~\ref{detect1} and Figure~\ref{AP_change}. When RFAConv is used to replace some convolution, the network achieved significantly improved detection results with only a small increase in the number of parameters and computational overhead. Compared to other attention, RFA still brings considerable benefits to the detection network. In experiments, we once again verify the effectiveness of RFCBAM and RFCA, which exhibits better performance of convolutional operation compared to the original CBAM and CA. This is precisely thanks to the consideration of receptive-field spatial features to solve the problem of convolutional parameter sharing. Time represents the total time spent processing an image during validation. It can be clearly seen that the model constructed with the novel convolutioanl operation has an increase in time for processing a image. Therefore, if real-time is pursued, the number of replacement convolutions should not be too many.

\begin{table*}[H]
	\centering
	\footnotesize
	\caption{Object detection $AP_{50}$, $AP_{75}$, $AP$, $AP_{S}$, $AP_{M}$, and $AP_{L}$ on the COCO2017 validation sets. We adopt the YOLOv5n, YOLOv7-tiny, and YOLOv8n detection framework and replace the original convolution with the novel convolutioanl operation constructed by attention mechanism.}
	\renewcommand\arraystretch{0.5}
	\small
	\setlength{\tabcolsep}{1mm}{
		\begin{tabular}{lccccccccc}
			\toprule
			Models & FLOPS(G) & Param(M) & $AP_{50}$(\%) & $AP_{75}$(\%) & AP(\%) & $AP_{S}$(\%)& $AP_{M}(\%)$& $AP_{L}(\%)$ & Time(ms)    \\
			\midrule
			YOLOv5n          & 4.5  & 1.8  & 45.6  & 28.9 & 27.5  & 13.5  & 31.5  & 35.9 & 4.4 \\
			+ CAMConv(r)     & 4.5  & 1.8  & 45.6  & 28.3 & 27.4  & 13.8  & 31.4  & 35.8 & 5.2 \\
			+ CBAMConv(r)    & 4.5  & 1.8  & 45.5  & 28.6 & 27.6  & 13.6  & 31.2  & 36.6 & 5.4 \\
			+ CAConv(r)      & 4.5  & 1.8  & 46.2  & 29.2 & 28.1  & 14.3  & 32.0  & 36.6 & 4.8 \\
			+ TAConv(r)      & 4.8  & 1.9  & 46.3  & 29.0 & 27.9  & 13.9  & 31.5  & 36.5 & 5.5 \\
			+ HAMConv(r)     & 4.6  & 1.9  & 45.7  & 28.2 & 27.4  & 15.0  & 31.3  & 35.6 & 5.7 \\
			+ MCAConv(r)     & 4.5  & 1.9  & 45.3  & 28.6 & 27.3  & 13.8  & 31.0  & 36.0 & 5.5 \\
			+ RFAConv(r)     & 4.7  & 1.9  &47.3           & 30.6         &29.0          &14.8   &33.4            &37.4           &5.3\\
			+ RFCBAMConv(r)  & 4.7  & 1.9  &\textbf{48.2}  & 30.8         & 29.4         & 15.2  & \textbf{33.9}  & 38.0          & 5.4 \\
			+ RFCAConv(r)    & 4.8  & 1.9  & 48.0          &\textbf{30.9} &\textbf{29.5} & \textbf{15.6}  & 33.3           & \textbf{38.2} & 5.5 \\
			\midrule
			YOLOv7-tiny      & 13.7 & 6.2   &53.8   &38.3  &35.9   & 19.9  & 39.4  & 48.8 & 3.4 \\
			+ TAConv(r)      & 13.9 & 6.2   &54.6   &39.6  &36.7   & 20.5  & 41.1  & 49.1 & 4.1 \\
			+ HAMConv(r)     & 13.7 & 6.2   &53.8   &38.4  &35.9   & 20.4  & 39.6  & 48.8 & 4.2 \\
			+ MCAConv(r)     & 13.7 & 6.2   &53.9   &38.9  &36.2   & 20.2  & 40.1  & 48.8 & 3.9 \\
			+ RFAConv(r)     & 14.1 & 6.3   &55.1          &\textbf{40.1}  &\textbf{37.1}   &\textbf{20.9} &\textbf{41.1} &50            &4.4\\
			+ RFCBAMConv(r)  & 14.1 & 6.3   &\textbf{55.2} & 40.0          & 37.1           & 20.1         & 41.1         &\textbf{50.4} & 4.6 \\
			+ RFCAConv(r)    & 14.2 & 6.3   & 55.1         & 40.0          & 37.1           & 20.8         & 41.0         & 49.5         & 4.6 \\
			\midrule
			YOLOv8n          & 8.7  & 3.1  &51.9   &39.7  &36.4   &18.4   &40.1   & 52.0 & 4.2 \\
			+ CAMConv(r)     & 8.8  & 3.1  &51.6   &39.0   &36.2  &18     &39.9   & 51.2 & 4.5 \\
			+ CBAMConv(r)    & 8.8  & 3.1  &51.5   &39.6  &36.3   &18.3   &40.1   & 51.5 & 4.6 \\
			+ CAConv(r)      & 8.8  & 3.1  &52.1   &39.9  &36.7   &17.8   &40.3   & 51.6 & 4.3 \\
			+ TAConv(r)      & 9.0  & 3.2  &52.0   &39.6  &36.5   &18.0   &39.8   & 51.2 & 4.8 \\
			+ MCAConv(r)     & 8.8  & 3.2  &51.6   &39.5  &36.3   &18.0   &40.0   & 51.1 & 4.5 \\
			+ RFAConv(r)     & 9.0  & 3.2  &53.4         &41.1           &37.7           &18.9          &41.8          &52.7          & 4.5 \\
			+ RFCBAMConv(r)  & 9.0  & 3.2  &53.3         &40.9            & 37.5         &19.0          &41.6          & 52.5          & 4.8 \\
			+ RFCAConv(r)    & 9.1  & 3.2  &\textbf{53.9} &\textbf{41.7}  &\textbf{38.2} &\textbf{19.7} &\textbf{42.3} &\textbf{53.5} &4.7\\
			\bottomrule
	\end{tabular}}
	\label{detect1}
\end{table*}

\subsection{Object detection experiments on VOC7+12 and Roboflow-100}
In order to validate our method again, we select the VOC 7+12 dataset for experiments. VOC 7+12 is a mixture of VOC2007 and VOC2012, with a total of 16551 training sets and 4952 verification sets. Similar to the experiments on COCO2017, we conduct experiments on advanced detection models such as YOLOv5n, YOLOv5s, YOLOv7-tiny, and YOLOv8n. All hyperparameter settings and network structure are the same as in the previous section. 
\begin{table}[h]
	\centering
	\footnotesize
	\caption{Object detection mAP50 and mAP on the VOC7+12 validation set.}
	\renewcommand\arraystretch{0.5}
	\setlength{\tabcolsep}{1mm}{
		\begin{tabular}{lcccc}
			\toprule
			Models & FLOPS (G) & Param (M)  & mAP (\%) &Time (ms)\\
			\midrule
			YOLOv5n          & 4.2    & 1.7    & 41.5         & 2.7 \\
			+ CAMConv(r)     & 4.2    & 1.7    & 41.4         & 2.9 \\
			+ CBAMConv(r)    & 4.3    & 1.7    & 41.9         & 3.0 \\
			+ CAConv(r)      & 4.3    & 1.7    & 42.4         & 3.0 \\
			+ TAConv(r)      & 4.5    & 1.8    & 42.5         & 2.8 \\
			+ MCAConv(r)     & 4.2    & 1.8    & 41.1         & 2.9   \\
			+ HAMConv(r)     & 4.3    & 1.8    & 41.3         & 2.9   \\
			+ RFAConv(r)     & 4.5    & 1.8    &\textbf{43.3} & 3.0 \\
			\midrule
			YOLOv5s          & 15.9   & 7.1    & 48.9         & 3.0\\
			+ CAMConv(r)     & 16.0   & 7.1    & 48.5         & 3.5 \\
			+ CBAMConv(r)    & 16.0   & 7.1    & 49.0         & 3.7 \\
			+ CAConv(r)      & 16.1   & 7.1    & 49.6         & 3.1 \\
			+ TAConv(r)      & 16.5   & 7.1    & 50.3         & 3.8 \\
			+ HAMConv(r)     & 16.0   & 7.1    & 48.5         & 4.3   \\
			+ MCAConv(r)     & 16.0   & 7.1    & 49.2         & 3.7   \\
			+ RFAConv(r)     & 16.4   & 7.2    & 50.0         & 5.1 \\
			+ RFCBAMConv(r)  & 16.4   & 7.2    & 50.1         & 3.9 \\
			+ RFCAConv(r)    & 16.6   & 7.2    &\textbf{51}   & 4.4\\
			\midrule
			YOLOv7-tiny      & 13.2   & 6.1    & 50.2         & 2.5 \\
			+ CAMConv(r)     & 13.2   & 6.1    & 50.3         & 2.7 \\
			+ CBAMConv(r)    & 13.2   & 6.1    & 50.1         & 2.7 \\
			+ CAConv(r)      & 13.2   & 6.1    & 50.5         & 2.7 \\
			+ TAConv(r)      & 13.4   & 6.1    & 50.4         & 2.8 \\
			+ MCAConv(r)     & 13.2   & 6.1    & 50.2         & 2.8 \\
			+ HAMConv(r)     & 13.2   & 6.1    & 50.1         & 2.7 \\
			+ RFAConv(r)     & 13.6   & 6.1    &\textbf{50.6} & 3.8 \\
			\midrule
			YOLOv8n          & 8.1   & 3.0     & 53.5         & 3.0 \\
			+ CAMConv(r)     & 8.1   & 3.0     & 52.8         & 3.1 \\
			+ CBAMConv(r)    & 8.2   & 3.0     & 53.3         & 3.1 \\
			+ CAConv(r)      & 8.2   & 3.0     & 53.8         & 2.9 \\
			+ TAConv(r)      & 8.4   & 3.0     & 54.0         & 3.3 \\
			+ MCAConv(r)     & 8.1   & 3.0     & 52.6         & 3.1 \\
			+ RFAConv(r)     & 8.4   & 3.0     &\textbf{54.0} & 3.2 \\
			\bottomrule			
	\end{tabular}}
	\label{detect2}
\end{table}
Following most of the work, we also report mAP. As shown in Table~\ref{detect2}. As with the previously obtained conclusions, in all experiments, after we replace some of the convolution operations in the network using RFAConv, the network gained significant improvement by adding only a small number of parameters and computational overhead. Meanwhile, RFA obtains an outstanding performance compared to other attentions. Moreover, in some experiments, we similarly experiment with networks constructed by RFCBAMConv and RFCAConv. The results again validate their advantages. For a fair comparison of other attention work, TAConv, HAMConv and MCAConv are still compared in advanced detection networks. All the experiments demonstrate that our proposed method obtains better performance.

Compared to the VOC7+12 dataset, Roboflow-100 includes 100 datasets from different domains. Therefore, to further demonstrate the effectiveness and generalizability of the proposed method, we conduct corresponding experiments on the large-scale Roboflow-100 dataset. We select YOLOv5n as the baseline network, training each network for 100 epochs, and the final reported results are obtained by averaging the experimental results from the 100 datasets. The experimental results are shown in Table~\ref{detect3}. Compared to other methods, our proposed method achieves the best detection accuracy by replacing the convolutions in YOLOv5. The object detection experiments on VOC7+12 and Roboflow-100 further demonstrate the good generalization of our method.

\begin{table*}[h]
	\centering
	\footnotesize
	\caption{Object detection experiment results based on the Roboflow-100 dataset.}
	\renewcommand\arraystretch{0.5}
	\setlength{\tabcolsep}{1mm}{
		\begin{tabular}{lcccccc}
			\toprule
			Models           & FLOPS (G) & Param (M) &Precision (\%) & Recall (\%)  & mAP50 (\%) &mAP (\%)\\
			\midrule
			YOLOv5n          & 4.2    & 1.7    & 64.85        & 59.09  & 56.52  & 34.28 \\
			+ CAMConv(r)     & 4.2    & 1.7    & 64.01        & 59.89  & 56.53  & 34.44  \\
			+ CBAMConv(r)    & 4.3    & 1.7    & 64.55        & 58.92  & 56.4  & 34.6\\
			+ CAConv(r)      & 4.3    & 1.7    & 65.01        & 59.5   & 57.01  & 34.81 \\
			+ TAConv(r)      & 4.5    & 1.8    & 63.99        & 58.25  & 56.23  & 33.99\\
			+ MCAConv(r)     & 4.2    & 1.8    & 64.8         & 59.6  & 56.57  & 34.35 \\
			+ HAMConv(r)     & 4.3    & 1.8    & 65.01        & 59.93  & 56.84  & 34.67\\
			+ RFAConv(r)     & 4.5    & 1.8    & 65.32        & 60.15  & 57.24  & 35.16\\
			+ RFCBAMConv(r)  & 4.5    & 1.8    & \textbf{65.52}        & \textbf{60.42}  & \textbf{57.91}  & \textbf{35.56}\\
			+ RFCAonv(r)     & 4.6    & 1.8    &65.44 & 60.31 & 57.85 & 35.37 \\
			\bottomrule			
	\end{tabular}}
	\label{detect3}
\end{table*}

\subsection{Semantic segmentation experiments on VOC2012}

In order to validate the advantages of our method again, we conduct semantic segmentation experiments on the VOC2012 dataset, selecting DeepLabplusV3 \cite{chen2018encoder} and the backbone network ResNet18 to conduct related experiments. Since the pre-trained weights of the backbone network are not available for other attention works, we only train a few networks that are able to take the pre-trained weights of the backbone network in the ImageNet-1k experiments. We report the results for outputs at two different step sizes, 8 and 16, respectively.

\begin{table}[h]
	\centering
	\footnotesize
	\caption{Results of experiments comparing different the novel convolutional operation based on DeepLabPlusV3.}
	\renewcommand\arraystretch{0.5}
	\setlength{\tabcolsep}{6mm}{
		\begin{tabular}{lcc}
			\toprule
			Backbone  & Stride   & MIOU (\%)    \\
			\midrule
			ResNet18        & 8       &58.9  \\
			+ CAMConv(r)    & 8       &60.9  \\
			+ CBAMConv(r)   & 8       &59.3  \\
			+ CAConv(r)     & 8       &62.1  \\
			+ RFAConv(r)    & 8       &60.8  \\
			+ RFCBAMConv(r) & 8       &62.1  \\
			+ RFCAConv(r)   & 8       &\textbf{63.9} \\
			\midrule
			ResNet18        & 16       &64.6  \\
			+ CAMConv(r)    & 16       &65.5  \\
			+ CBAMConv(r)   & 16       &63.6  \\
			+ CAConv(r)     & 16       &66.6  \\
			+ RFAConv(r)    & 16       &65.4  \\
			+ RFCBAMConv(r) & 16       &67.7  \\
			+ RFCAConv(r)   & 16       &\textbf{68.0} \\
			\bottomrule
	\end{tabular}}
	\label{Semantic1}
\end{table}

\begin{figure*}[h]
	\centering
	\includegraphics[trim=0 0 0 0,clip,scale=0.85]{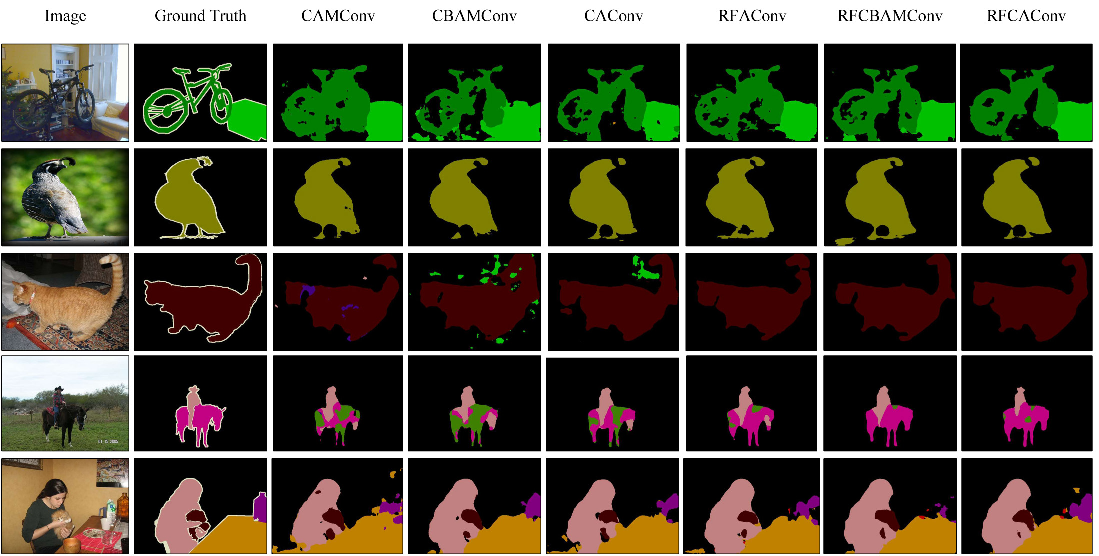}
	\caption{Segmentation visualisation results of ResNet18 constructed with different convolutions. It is evident that RFAConv and RFCBAMConv allow the network to segment the detailed contours of the target more clearly than the other methods. CAConv differs from RFCAConv in that the latter focuses on the receptive-field spatial features, which proves the importance of the receptive-field spatial features.}
	\label{semantic-visual}
\end{figure*}

As shown in Table~\ref{Semantic1}, we found that the semantic segmentation network constructed by RFAConv achieved better results than the original model, but compared to CAConv, CAMConv, the performance of RFAConv is not good. After thinking, we assert that RFAConv lacks consideration for long-distance information, while semantic segmentation tasks rely on long-distance information. CAConv, CAMConv and CBAMConv capture long range information by global averaging pooling to obtain global information. Although CBAMConv produce poor results for semantic segmentation, the improved RFCBAM obtain the good performance. This again demonstrates that spatial attention can again improve network performance through our approach, which simply requires making spatial attention attend to the receptive-field spatial feature. To further underscore the merits of our proposed approach, we randomly select 5 images from the VOC2012 dataset and input them into the ResNet18 model constructed with various convolution operations for segmentation visualization and analysis. As illustrated in Figure \ref{semantic-visual}, it is evident that the networks built using RFCBAMConv and RFCAConv, excel in delineating the intricate contours of the target compared to other methods. As previously noted, the significance of the receptive-field spatial features is paramount, as highlighted in the segmentation visualization contrast between CAConv and RFCAConv. The distinguishing factor of RFCAConv, in contrast to CAConv, lies in its emphasis on leveraging the receptive-field spatial features to tackle convolution parameter sharing challenges. Consequently, RFCAConv demonstrates superior segmentation outcomes when compared to CAConv.

\subsection{Discussions}

In classification experiments, RFAConv demonstrates substantial accuracy improvements over baseline models employing standard convolutions, despite marginal increases in parameters and computational overhead, as quantified in Table~\ref{classification1}. 
This performance gap stems from fundamental architectural differences: while standard convolutions apply static kernel parameters across all spatial positions, RFAConv implements location-adaptive parameter modulation through learned attention weights for each receptive-field slide. This dynamic mechanism enables superior adaptation to spatial variations in target scale and distribution, addressing critical limitations of conventional spatial attention approaches. Existing spatial attention methods exhibit suboptimal performance due to their non-discriminative weight allocation across receptive-field regions, as evidenced by overlapping attention map. Such uniform weighting fails to resolve the parameter-sharing constraints inherent to large convolutional kernels. For 3 $\times$ 3 convolutions, RFAConv outperforms ResNet18 and ResNet34 models constructed using other attention mechanisms, a finding supported by the results in Table~\ref{classification1}. Unlike convolution operations constructed by other attention mechanisms, RFAConv assigns unique attention weights to each receptive-field slider, enabling the network to highlight key targets, as depicted in Figure~\ref{Grad-CAM1} for visualization results using the CAM algorithm. Furthermore, enhancing the focus of CA and CBAM on receptive-field spatial features contributes to improved performance. Extending this paradigm, we develop RFCBAM and RFCA by integrating receptive-field spatial feature enhancement into existing attention frameworks, as evidenced in Table~\ref{classification2}, where RFCBAMConv and RFCAConv notably improve the accuracy of ResNet18. In comparison to networks built with CBAMConv and CAConv in Table~\ref{classification1}, networks constructed by RFCBAMConv and RFCAConv effectively optimize network performance, underscoring the importance of the receptive-field spatial features. Figure~\ref{Grad-CAM2} further illustrates that RFCBAMConv and RFCAConv emphasize key objects more effectively than CAConv and CBAMConv by adapting different attentional weights for each receptive-field slider.

Experimental validation demonstrates that our proposed method consistently enhances detection performance across benchmark datasets, with quantitative improvements substantiated in Tables~\ref{detect1} and Table~\ref{detect2}. On the COCO2017 dataset, the framework achieves significant gains in average precision compared to baseline models, attributed to its capacity to learn location-specific attentional weights for the receptive-field attention method.  This design effectively captures spatial variations in target characteristics across complex scenes. However, performance improvements on the VOC dataset remain comparatively limited, likely due to the dataset’s smaller scale and reduced inter-target variability in simpler images. The proposed method learns distinct attentional weights for each receptive-field slider to account for variations in target information at different locations. As a result, the proposed methods demonstrate minimal improvement on VOC data, whereas it significantly enhances network performance on COCO2017.

Empirical observations indicate that while RFAConv offers distinct advantages in semantic segmentation tasks, its performance is not the best. Compared to contemporary attention-enhanced convolutional methods, RFAConv demonstrates measurable improvements over conventional convolution operations, yet its quantitative metrics remain below those of leading attention-based counterparts. This can be explained by the fact that the accuracy of semantic segmentation tasks heavily relies on long-distance information \cite{hou2021coordinate}, which the design of RFA neglects to capture in spatial terms, leading to its less outstanding performance in semantic segmentation tasks. In contrast, CA performs global pooling in both the H and W directions, thereby partially considering long-distance information and achieving good performance. Although CBAM and CAM also consider long-distance information through global pooling, their compression methods result in significant information loss, leading to suboptimal performance \cite{hou2021coordinate}. To further demonstrate the importance of receptive-field spatial features in semantic segmentation tasks, we conduct experiments with RFCBAMConv and RFAConv. The experimental results again demonstrate a significant performance improvement by focusing the attention of existing attention mechanisms on receptive-field spatial features. For example, the comparison between CAConv and RFCAConv highlights this fact.

In summary, the collective experimental findings unequivocally establish RFAConv as a robust solution capable of surpassing standard convolution operations to boost network performance, and only adds a small amount of computational overhead, number of parameters and inference time. Its plug-and-play compatibility enables direct integration into existing architectures through layer-wise substitution of standard convolutions, requiring no structural redesign. Notably, RFAConv assigns unique attention weights to the receptive-field sliders at each position, resulting in significant performance enhancements for datasets with substantial disparities in positional information. Thus, for datasets exhibiting significant discrepancies in location information, such as UAV images, underwater images, and medical images where target characteristics vary across different locations, RFAConv excels in capturing intricate details, thereby enhancing network accuracy.\\

\noindent \textbf{Limitations.} To address the problem of standard convolutional parameter sharing, corresponding attentional weights are learned for each position of the receptive-field slider, which can lead to a larger memory overhead for the network. Therefore devices with limited storage resources may be limited in using our method to construct 3 $\times$ 3 non-shared parameter convolutions. However, it is entirely possible to learn 2 $\times$ 2 attentional weights for each position, which does not completely solve the problem of parameter sharing of standard 3 $\times$ 3 convolution kernels, but can alleviate the problem of parameter sharing of standard convolutions to some extent. This approach improves network performance while maintaining a small increase in memory overhead. Another perfect solution is to build non-square convolutional kernel operations to further flexibly adjust the memory overhead and build non-shared parameter convolutions to boost network performance.

\section{Conclusion}

Existing works predominantly concentrate on leveraging spatial attention to improve network performance, exploring its contributions to feature extraction, information selection. Although they indeed show the effectiveness of spatial attention to some degree, they tend to analyze spatial attention from a conventional viewpoint, overlooking the deeper relationship between its intrinsic mechanism and convolutional kernels. This work aims to showcase the efficacy of spatial attention by dissecting the inherent link between spatial attention and the standard convolution from a fresh perspective. It essentially solves the problem of convolution kernel parameter sharing. However, the current performance of spatial attention mechanisms with large convolution kernels remains restricted. Due to the shared attention weights within each receptive-field, the issue of parameter sharing in convolutional kernels is not entirely resolved. 

To address this limitation, a novel attention mechanism called Receptive-Field Attention (RFA) is proposed, which emphasizes the receptive-field spatial feature and the significance of each feature within the sliding window. Building on the RFA concept, a Receptive-Field Attention Convolution (RFAConv) is developed to substitute the traditional standard convolution operation. It significantly improves network performance with almost negligible computational overhead and parameter increase, and is expected to replace traditional standard convolutional operations as a core component of neural network architectures for classification, target detection and semantic segmentation tasks. Furthermore, this work underscores the importance of spatial attention mechanisms focusing on receptive-field spatial features to further enhance performance. Therefore, we introduce improved versions of RFCBAM and RFCA based on CBAM and CA, and combine them with standard convolutional operations to design non-shared parameter convolutional operations. Subsequently, we meticulously conduct a large number of classification, object detection, and semantic segmentation experiments on authoritative datasets like ImageNet, COCO, VOC, and Roboflow to thoroughly validate their efficacy and refinement.

Although the proposed method significantly enhances network performance, it requires learning distinct attentional weights for each position of the receptive-field slider to tackle the standard convolutional parameter sharing issue. Consequently, this leads to an increase in the network's memory overhead. A perfect solution is to build non-square convolutional kernel operations to further flexibly adjust the memory overhead and build non-shared parameter convolutions to boost network performance. Therefore, In the future, we will develop a convolution operation with an arbitrarily shaped convolution kernel to flexibly tune its corresponding memory overhead. Moreover, we anticipate a shift towards increased research on receptive-field spatial features, aiming to leverage the benefits of spatial attention for addressing parameter sharing challenges in standard convolutional operations to dramatically improve network performance.

\section*{CRediT authorship contribution statement}
\noindent\textbf{Xin Zhang:} Conceptualization, Methodology, Validation, Software, Formal analysis, Investigation, Writing-original draft, Writing-review \& editing, Visualization.
\textbf{Chen Liu:} Resources, Data Curation, Writing-review \& editing, Validation.
\textbf{Tingting Song:} Methodology, Resources, Writing-review \& editing, Project administration, Supervision, Funding acquisition. 
\textbf{Degang Yang:} Methodology, Resources, Writing-review \& editing, Project administration, Supervision, Funding acquisition.
\textbf{Yichen Ye:} Data Curation, Resources, Formal analysis, Writing-review \& editing.
\textbf{Ke Li:} Data Curation, Visualization, Resources, Formal analysis.
\textbf{Yingze Song:} Data Curation, Visualization, Resources, Formal analysis.

\section*{Declaration of Competing Interest}
The authors declare that they have no known competing financial interests or personal relationships that could have appeared to influence the work reported in this paper.

\section*{Acknowledgments}

This research was supported in part by National Natural Science Foundation of China (62441506), in part by Natural
Science Foundation of Chongqing (CSTB2022NSCQ-MSX1200), in part by the Science and Technology Research Program of Chongqing
Municipal Education Commission (KJQN202200537 and KJZD-M202300502), in part by Chongqing Normal University PhD Start-up Fund
(21XLB035), and in part by Chongqing Normal University Postgraduate Research and Innovation Program (YKC24017).

\bibliographystyle{elsarticle-num}
\bibliography{reference}

\end{document}